\pdfoutput=1

\documentclass[11pt]{article}

\usepackage[]{acl}

\usepackage{acl}

\usepackage{times}
\usepackage{latexsym}
\usepackage{amssymb}
\usepackage{xcolor}
\usepackage{color,colortbl}
\usepackage{amsmath}
\usepackage{multirow}
\usepackage{graphicx}
\usepackage{paralist}
\usepackage{booktabs}
\usepackage{array}
\usepackage{multicol}
\usepackage{subcaption}
\usepackage{enumitem}
\usepackage{tabularx}
\usepackage{color, colortbl}
\usepackage{booktabs} 
\usepackage{mdframed}

\definecolor{Gray}{gray}{0.9}
\usepackage{hyperref}

\usepackage{algorithm,algpseudocode}
\usepackage{newfloat}
\usepackage{listings}

\definecolor{Gray}{gray}{0.9}

\usepackage[T1]{fontenc}

\usepackage[utf8]{inputenc}

\usepackage{microtype}

\title{Enhancing Post-Hoc Attributions in Long Document Comprehension\\via Coarse Grained Answer Decomposition}

\author {
    Pritika Ramu, Koustava Goswami, Apoorv Saxena, Balaji Vasan Srinivasan\\
    Adobe Research, India\\
    \small{\tt \{pramu, koustavag, apoorvs, balsrini\}@adobe.com}
}

\begin{document}

\maketitle


\begin{abstract}
Accurately attributing answer text to its source document is crucial for developing a reliable question-answering system. However, attribution for long documents remains largely unexplored. Post-hoc attribution systems are designed to map answer text back to the source document, yet the granularity of this mapping has not been addressed. Furthermore, a critical question arises: What exactly should be attributed? This involves identifying the specific information units within an answer that require grounding. In this paper, we propose and investigate a novel approach to the factual decomposition of generated answers for attribution, employing template-based in-context learning. To accomplish this, we utilize the question and integrate negative sampling during few-shot in-context learning for decomposition. This approach enhances the semantic understanding of both abstractive and extractive answers. We examine the impact of answer decomposition by providing a thorough examination of various attribution approaches, ranging from retrieval-based techniques to LLM-based attributors.
\end{abstract}
\section{Introduction}

The rise of Large Language Models (LLMs) and GenAI-based technologies has greatly increased their usability across various sectors, notably in grounded question-answering systems. However, to establish trust, it's crucial to attribute information obtained from source documents, especially given the tendency of these models to generate texts from their own knowledge bases \cite{huang2023survey}. With opaque LLMs like ChatGPT, there's a need to explore post-hoc attribution methods to enhance reliability. Targeted attribution is necessary in chat-based question-answering systems to improve user experience, as not every part of an answer requires attribution.

\begin{figure}[t!]
    \centering
    \includegraphics[width=\columnwidth]{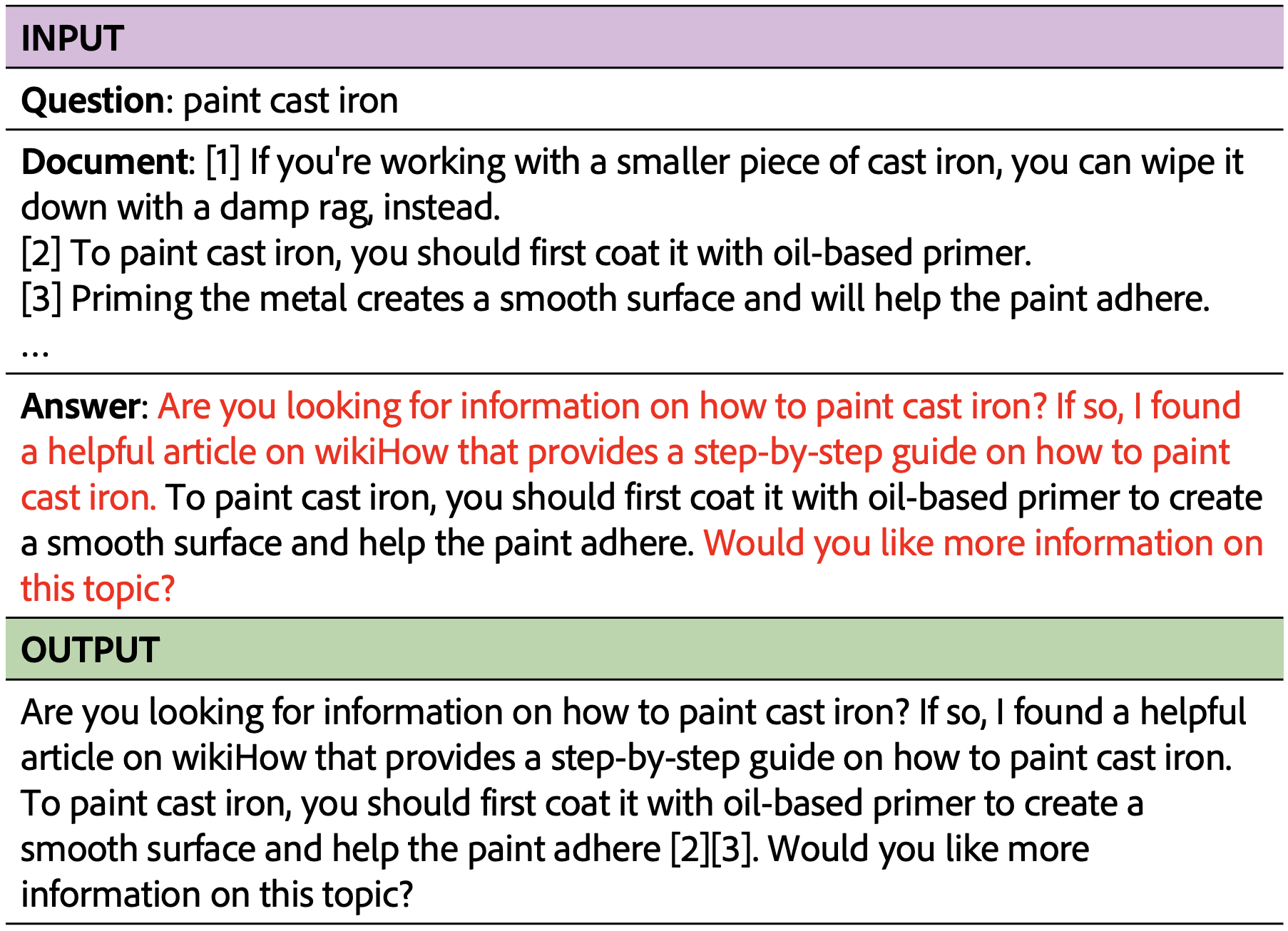}
    \caption{An example from Verifiability dataset. The input to the post-hoc attribution system is the question, document and answer. The output is evidence sentences from the document. Text marked in \textcolor{red}{red} do not require attribution.}
    \label{fig:proposed}
\end{figure}

Prior research addresses generating attributions alongside responses in open domains \citep{gao2023enabling,gao2023rarr}, either per sentence or per paragraph \cite{bohnet2023attributed}. However, attributing responses in long document sequences is challenging, and inline attribution falls short \cite{gao2023enabling}. Additionally, determining \textit{when to cite} is crucial, as inappropriate or excessive citations can lead to redundancy \cite{huang-chang-2024-citation}. Moreover, excessive in-line citations can diminish creativity in generated content \cite{huang-chang-2024-citation}. \cite{guo-etal-2022-survey} explores fine-tuning to enhance attributions, but these models are domain-specific, limiting their adaptability.

 With rise of LLMs, the attribution task is viewed as a retrieval and mapping task \cite{li-etal-2022-encoder}. This shift presents challenges, especially in abstractive question-answering scenarios, where retrieving context is not straightforward, and semantic relatedness must often be inferred from the entire document context. Notably, \cite{gao2023rarr} has underscored the limitations of retriever-based attribution engines, particularly in handling out-of-distribution knowledge during context processing. 

Despite these efforts, existing methods treat answers as singular attributable elements and strive to map them back to long sequence contexts. However, answers can encompass multiple facts and are contextually dependent on the posed question. Existing fine-tuning or retriever-based systems face challenges in identifying the specific attributable components within an answer. In this paper, we formulate post-hoc attribution as the task of identifying source sentences from a document that support \textit{attributable} parts of an answer when our input consists of a question, answer, and the document from which the answer is obtained.

\textbf{Firstly}, we propose using template-based in-context learning as a method to achieve question-contextualized decomposition of answers. This technique helps break down answers that aligns with the question, which aids identifying references in the document context. This alignment is crucial as it enhances the identification of the specific information units necessary to answer the question effectively. Empirical observations suggest that guiding a language model for specific tasks in few-shot settings can lead to improved adaptability \cite{DBLP:journals/corr/abs-2303-13217}. Building upon this insight, we recognize the significance of guiding a language model for in-context decomposition. Additionally, we introduce \textit{negative sampling}-based in-context learning to enable language models to discern between good and bad decomposition.

\textbf{Secondly}, we analyze the use of different retrieval methods as attributors and compare their performance using four decomposition methods. We observe that, on average, question-contextualized coarse-grain decomposition results are comparable, and in some settings, better than using non-decomposed sentences. This serves as a regression test for our methodology without degrading retriever performance. Moreover, our experimental results on the Citation Verifiability dataset highlight that contextualized coarse-grain decomposition for retriever-based attributors (BM25, GTR, MonoT5) achieves, on average, a \textbf{3\%} gain in precision over baseline models, emphasizing the efficacy of decomposition for attributions (Table \ref{tab:retriever}).

\textbf{Thirdly,} we examine the use of LLMs as post-hoc attributors and the advantages they offer by considering the context of the question and the decomposed answer. We observe a significant improvement in performance both empirically and qualitatively when providing question-contextualized coarse-grain decomposition of answers to LLMs. By leveraging LLMs as attributors and incorporating question-contextualized coarse-grain decomposition, we achieve state-of-the-art performance on the QASPER and Verifiability datasets (Table \ref{tab:llm}). 
\section{Related Work}

\begin{figure*}[t]
  \centering
  \includegraphics[width=\textwidth]{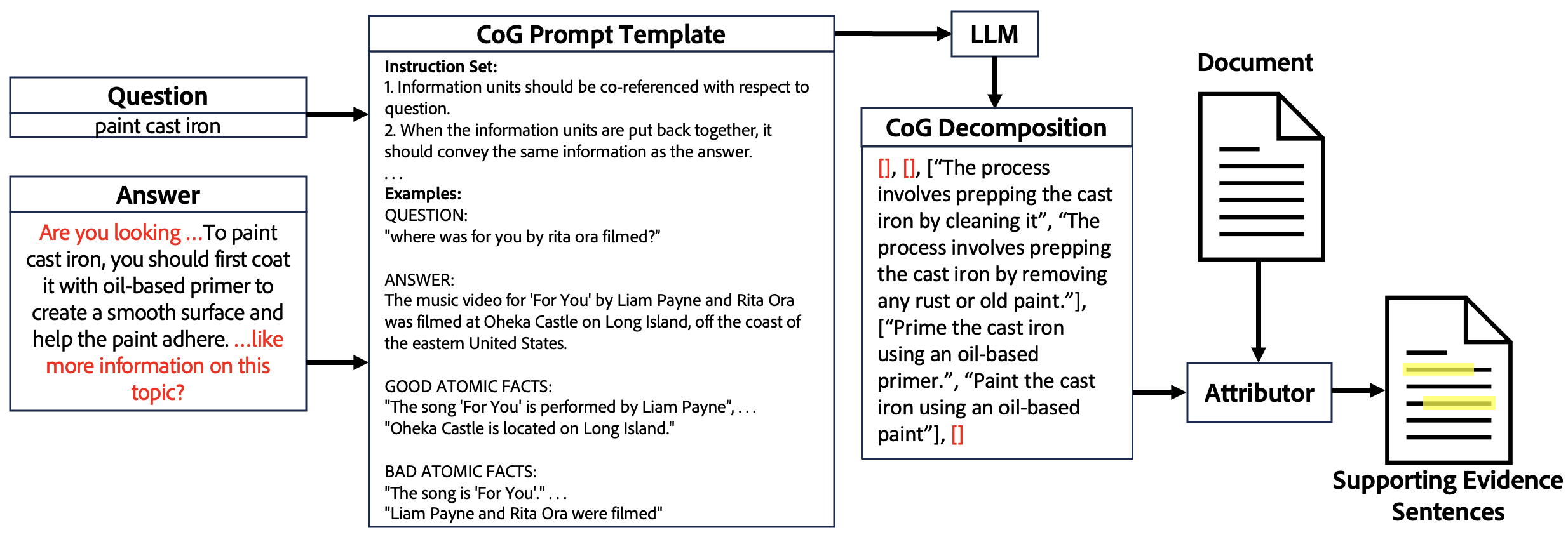}
  \caption{Pipeline for attribution: Answers are decomposed and sent to the attributor for identifying evidences.}
  \label{fig:pipeline}
\end{figure*}

Multiple works have established the need for ensuring a trustworthy model and how citations and attributions can help ensure that reliable information is provided or to detect if any information is hallucinated \cite{huang2023citation, litschko2023establishing, rashkin2022measuring, venkit2024confidently}. Efforts to create truthful AI have been highlighted, emphasizing the importance of fact-checking and claim verification, even in the pre-LLM era \cite{evans2021truthful}. Notably, \cite{petroni2022improving} focuses on improving the identification of claims from Wikipedia that lack support from citations, with a specific emphasis on identifying supporting evidence at a paragraph level. In addition to unimodal systems, multimodal systems, with the introduction of generative AI, are becoming increasingly important to attribute contextual parts and build trustworthy AI systems \cite{DBLP:journals/corr/abs-1711-06104,DBLP:journals/inffus/HolzingerMSP21,DBLP:conf/emnlp/ZhaoCWJLQDGLLJ23,phukan2024peering}.

In the context of dialogue systems requiring background knowledge, issues with spurious correlations have been acknowledged, necessitating a more robust method for identifying what can be attributed in an abstractive setting \cite{dziri-etal-2022-evaluating}. \cite{mei2023foveate} introduces the identification of missing information and providing appropriate attributions to mitigate potential dangers due to misinformation. Meanwhile, \cite{sarti2023quantifying, sarti-etal-2023-inseq} delve into the analysis of feature attribution and interpretability of language models. A unique task is undertaken by \cite{funkquist2023citebench}, which addresses the creation of inline citations for scientific articles from abstracts of other scientific articles.

\cite{sancheti-etal-2024-post} introduced fine-grained attribution of answers but did not cover fact-based attribution. The task of fact-based attribution—assigning facts from answers to a list of evidence sentences rather than paragraphs—is under-explored in the existing literature. This underscores the need for further exploration in fact-based attribution, particularly concerning sentence-level source attributions and the challenges posed by long document sequences in question-answering systems.
\section{Method}

We propose a 2-fold process: First, decompose the answer into smaller information units, then employ an attributor to map these units to evidence from the source document (Fig. \ref{fig:pipeline}). We investigate using retrieval methods and various LLMs for attribution.

\subsection{Task Definition}

Given a question $Q$, a collection of evidence sentences $E = e_1, ... , e_n$ extracted from document D, and an answer $A$ to question $Q$, the objective is to identify supporting evidence sentences (attributions) $e_i \in E$ for each answer part $a_i \in A = a_1,..., a_m$. Each $a_i$ may have multiple evidence sentences associated with it. An answer part is defined as a single complete sentence from an answer. 

\subsection{Answer Decomposition}

To link answer components to a given document, we decompose the answer parts into smaller units known as information units.

\subsubsection{Definition}

Answers to a question often extend over considerable lengths, encapsulating diverse facts and information. Within these responses, the veracity of facts can vary, presenting a mix of true and false statements. When tasked with extracting evidence from a document in response to a question, the determination of factual accuracy relies on whether a given fact is substantiated by at least one supporting evidence from the source document.

The granularity at which we deconstruct an answer can vary, spanning from fine-grained to coarse-grained representations. While previous studies have investigated the use of a sentence from an answer as an information unit, it is important to note that some sentences may be intricate, containing multiple conjunctions of information. This complexity underscores the need for a nuanced approach to information unit selection in order to capture what is to be attributed from an answer accurately. Identifying which information from the answer requires grounding is an important task to prevent grounding unnecessary sentences.

For answer part $a_i \in A$, we can decompose $a_i$ as ${iu_1,...,iu_n}$ such that $\text{information}(iu_i) \subseteq \text{information}(a_i)$.

\subsubsection{ Revisiting Fine Grained Decomposition}
FActScore \cite{min2023factscore} defines an atomic fact as a short sentence that conveys a single piece of information. This method assesses the factual precision of long texts through fine-grained decomposition but does not consider what specifically requires attribution and decomposition. By decomposing one sentence at a time, it often overlooks contextual semantics and the context of the question. To address these limitations, we propose Coarse Grained Decomposition (CoG).

\subsubsection{Coarse Grained Decomposition (CoG)}
Determining the appropriate granularity for decomposing an answer is subjective. For tasks requiring evidence extraction from a source document, it is impractical to set a strict upper limit on the number of facts per information unit. To effectively guide this process, we propose a question-contextualized decomposition approach. By developing a question-aware prompt template, we direct the LLM to generate information units that align closely with the specific context and requirements of the question. 

We design the prompt template for coarse grained decomposition by giving instructions on how an answer should be broken down and how not to be broken down. LLMs are known to perform poor on negation tasks \cite{truong-etal-2023-language}. So instead of prompting the model to not generate a poor information unit, we provide a template to generate good and bad information units enabling us to parse out the required good information units. We provide the question and entire answer to the decomposer to obtain a more contextualized decomposition. See Appendix \ref{sec:decomp_schema} for complete prompt.

\textbf{Instruction Schema.} The instruction schema is a generalizable schema that follows \cite{wang-etal-2022-super}. The \textbf{Task Definition} defines a given task and lists the input and output formats for the question, answer, and decomposition. The \textbf{Instruction Set} guides the language model to generate useful decompositions. We explain what good and bad instruction units are. Good information units are relevant, meaningful, and are directly associated with the question. They are broken down at logical conjunctions and, when reassembled, convey the same meaning as the original answer. Conversely, bad information units include redundant content, non-statements, or irrelevant information. \textbf{Positive Samples} illustrate the expected decomposition for a given question, aligning with the provided instructions. We include examples covering all instructions in the set. \textbf{Negative Samples} demonstrate negation instructions to the LLM, which typically struggles with negation tasks. These samples align with negation instructions.





\subsection{Classifier}
The classifier aids the attribution system's efficiency by deciding whether a sentence requires decomposition. This benefits in cost and latency by identifying and excluding simplistic sentences from decomposition. For instance, the sentence "Alex is an engineer." is already a simple sentence containing a single fact and therefore does not require decomposition. 

To systematically identify such cases, we have implemented a rule-based classifier that evaluates sentences based on their linguistic characteristics, specifically their part-of-speech (POS) tags, as described by Toutanova et al. \cite{toutanova-etal-2003-feature}. Let $S$ be the set of POS tags obtained from a sentence, with $N$, $P$, $V$, and $A$ representing Noun, Pronoun, Verb, and Article respectively. A sentence is deemed to not require decomposition if it satisfies the following constraint $(C_l)$:

\begin{align*}
    C_i = (|S\cap\{N\cup P\cup A\}| =|S|-1) \\
    C_j = (S-\{N\cup P\cup A\}\in V) \\
    C_k = (S \subseteq \{N\cup P \cup A\}) \\ 
    C_l = (C_i \land C_j) \lor C_k 
\end{align*}

This constraint implies that if a sentence consists solely of any combination of nouns, pronouns, or articles with at most one verb, it is classified as a simple sentence and is not decomposed\footnote{This constraint does not hold true if punctuation other than full stop (.) or quotes ("') exist.}. This classification rule acts as a high-precision filter, targeting sentences that are single independent clauses. These clauses are straightforward and typically do not benefit from further decomposition.

The design of this rule was based on linguistic rules and empirical observations to ensure high precision. While this rule is optimized for high precision, it is possible that simple sentences are not identified as simple sentences. In such instances, the integrity of the decomposition process is not compromised since the output remains effectively unchanged when processed by the decomposer.

\subsection{Attributors}
We evaluate two types of attributors for extracting evidence from documents: retrievers and LLMs. Retrievers rank evidence sentences from the document, while LLMs select the most appropriate evidences for each information unit.

\subsubsection{Retrievers}
We investigate various retrieval methods that serve as attributors by ranking evidence sentences within a document. Each information unit is treated as a query to retrieve evidence. For answer parts containing multiple information units, our goal is to select the most relevant evidence set. To optimize this selection, we use a greedy merging strategy.

Alogrithm \ref{alg:naive_greedy} shows a greedy merging algorithm on how the evidences are chosen for an answer part. Let $IU$ be the list of information units for an answer part, $E$ the list of evidences from the document and $L$ the final ordered list of evidences. $\text{score}(iu,e)$ refers to the score obtained from the retriever when $iu$ is the query and $e$ is a single evidence from the document. We rank the list of evidences based on $\text{score}(iu,e)$. If two information units have a high score for the same evidence sentence, we include the evidence only once. The intuition to using a greedy algorithm is to is to obtain only top evidences for an answer part. If an information unit has a low score with the evidences, it means that information is not attributable to the document.

\begin{algorithm}
\caption{Merging of Evidences for Answer Part}
\label{alg:naive_greedy}
\begin{algorithmic}[0]
\small
    \State $IU = [iu_1, iu_2, \ldots, iu_n]$
    \State $E = [[e_{11}, \ldots, e_{1m}], \ldots, [e_{n1}, \ldots, e_{nm}]] $
    \State Initialize an empty list $L$
    \For{each $iu$ in $IU$}
        \State Initialize a variable $\text{max\_score} \leftarrow -\infty$
        \State Initialize a variable $\text{best\_evidence} \leftarrow \text{null}$
        \For{each $e$ in $E[iu]$}
            \If{$\text{score}(iu,e) > \text{max\_score}$ and $e$ not in $L$}
                \State $\text{max\_score} \leftarrow \text{score}(iu,e)$
                \State $\text{best\_evidence} \leftarrow e$
            \EndIf
        \EndFor
        \If{$\text{best\_evidence} \neq \text{null}$}
            \State $L \leftarrow (\text{max\_score}, \text{best\_evidence})$
        \EndIf
    \EndFor
    \State Sort $L$ in descending order based on the score
\end{algorithmic}
\end{algorithm}



\subsubsection{Large Language Models}
Prior works have shown the ability of LLMs to generate text with citations, which reduces the hallucinations \cite{gao2023enabling}. We explore LLMs as attributors when the task is broken down into finding evidence for smaller sentences. We ask the LLM to find evidences that support information units for a given question. 

\textbf{Instruction Schema.} To guide the LLM in finding attribution for a question and its corresponding information unit, the instruction schema for using LLM as an attributor comprises two main components. The \textbf{Task Definition} defines the task of attributing an information unit from a list of retrieved sentences. The \textbf{Instruction Set} instructs the LLM to identify and select a valid list of evidence and sort such that the most relevant evidence appears first. For complete prompt, see Appendix \ref{sec:llm_att}.

\section{Datasets} \label{subsec:dataset}

We reformulate the Citation Verifiability dataset \cite{liu2023evaluating} and QASPER dataset \cite{dasigi2021dataset} for our task. The statistics of the dataset are present in Table \ref{tab:stats} (Appendix \ref{sec:data_stat}).
\subsection{Citation Verifiability Dataset}
 The Citation Verifiability dataset~\citep{liu2023evaluating} consists of questions from NaturalQuestions~\citep{kwiatkowski-etal-2019-natural} and ELI5~\citep{fan-etal-2019-eli5}. Answers are generated from search engines such as Bing Chat and NeevaAI. The answers contain inline citations pointing to web pages. Human annotators judged citations as \textit{fully}, \textit{partially}, or \textit{not} supporting sentences. For fully supported sentences, annotators provided supporting sentences from cited web pages. We consider these supporting sentences as attributions for an answer part. This dataset provides evidences on a sentence level and provides citations only for the informative parts of the answer.

\subsection{QASPER Dataset}
QASPER ~\citep{dasigi2021dataset} is a dataset for information-seeking questions and answers grounded in research papers. Each question is written by an NLP practitioner who only read the title and abstract of the corresponding paper, and the question seeks information present in the full text. The questions are then answered by a separate set of NLP practitioners who also provide supporting evidence for their answers. The answers in the dataset can be unanswerable, extractive (spans in the paper serving as the answer), or free-form. We take the evidences marked as supporting evidence (evidences with high agreement amongst annotators) as the ground truth attribution. This dataset does not provide the evidence on a sentence level, but on an answer level. 

\begin{table*}[t]
    \centering
    \scalebox{0.72}{
\begin{tabular}{c|c|c|c|c|c|c}
\toprule \multirow{2}{*}{ } & \multicolumn{3}{|c|}{ Verifiability } & \multicolumn{3}{|c}{ QASPER } \\
\midrule Decomposer+Attributor & Top 1 P/R/F1 & Top 2 P/R/F1 & Top 4 P/R/F1 & Top 1 P/R/F1 & Top 2 P/R/F1 & Top 4 P/R/F1 \\
\midrule  NIL+BM25 & $0.66 / \textbf{0.53} / 0.59$ & $0.44 / \textbf{0.65} / \textbf{0.53}$ & $0.27 / \textbf{0.72} / 0.39$ & $\textbf{0.42} / \textbf{0.18} / \textbf{0.26}$ & $\textbf{0.32} / \textbf{0.25} / \textbf{0.28}$ & $\textbf{0.22} / \textbf{0.31} / \textbf{0.25}$ \\
 FActScore+BM25 & $0.31 / 0.24 / 0.27$ & $0.24 / 0.34 / 0.28$ & $0.18 / 0.47 / 0.26$ & $0.24 / 0.09 / 0.13$ & $0.18 / 0.13 / 0.15$ & $0.11 / 0.16 / 0.13$ \\
(CoG - neg.)+BM25 & $0.67 / 0.51 / 0.59$ & $0.45 / 0.61 / 0.52$ & $0.28 / 0.69 / 0.40$ & $0.35 / 0.17 / 0.22$ & $0.30 / 0.22 / 0.25$ & $0.21 / 0.29 / 0.23$ \\
 CoG+BM25 & $\textbf{0.69} / 0.52 / \textbf{0.60}$ & $\textbf{0.46} / 0.62 / \textbf{0.53}$ & $\textbf{0.29} / 0.70 / \textbf{0.41}$ & $0.37 / 0.17 / 0.23$ & $0.30 / 0.23 / 0.26$ & $\textbf{0.22} / 0.30 / 0.24$ \\
\midrule NIL+GTR & $0.66 / \textbf{0.51} / 0.57$ & $0.43 / \textbf{0.62} / \textbf{0.51}$ & $0.27 / \textbf{0.72} / 0.39$ & $\textbf{0.41} / \textbf{0.18} / \textbf{0.25}$ & $\textbf{0.31} / \textbf{0.24} / \textbf{0.27}$ & $\textbf{0.22} / \textbf{0.31} / \textbf{0.26}$ \\
 FActScore+GTR & $0.57 / 0.44 / 0.50$ & $0.38 / 0.56 / 0.45$ & $0.24 / 0.66 / 0.35$ & $0.28 / 0.10 / 0.15$ & $0.21 / 0.15 / 0.18$ & $0.15 / 0.22 / 0.18$ \\
 (CoG-neg.)+GTR & $0.67 / 0.50 / 0.57$ & $0.44 / 0.58 / 0.50$ & $0.27 / 0.70 / 0.40$ & $0.37 / 0.16 / 0.22$ & $0.29 / 0.21 / 0.25$ & $0.21 / 0.30 / 0.25$ \\
 CoG+GTR & $\textbf{0.69} / \textbf{0.51} / \textbf{0.59}$ & $\textbf{0.45} / 0.59 / \textbf{0.51}$ & $\textbf{0.29} / 0.71 / \textbf{0.42}$ & $0.39 / 0.17 / 0.24$ & $0.30 / 0.23 / 0.26$ & $\textbf{0.22} / \textbf{0.31} / \textbf{0.26}$ \\
 \midrule  NIL+MonoT5 & $0.70 / \textbf{0.54} / 0.61$ & $0.47 / 0.68 / 0.55$ & $0.28 / \textbf{0.76} / 0.41$ & $\textbf{0.47} / \textbf{0.21} / \textbf{0.29}$ & $0.34 / \textbf{0.28} / \textbf{0.31}$ & $0.24 / 0.35 / 0.28$ \\
  FActScore+MonoT5 & $0.62 / 0.50 / 0.55$ & $0.41 / 0.61 / 0.49$ & $0.24 / 0.66 / 0.36$ & $0.39 / 0.16 / 0.23$ & $0.26 / 0.21 / 0.24$ & $0.18 / 0.26 / 0.21$ \\
   (CoG – neg.)+MonoT5 & $0.70 / 0.53 / 0.60$ & $0.46 / 0.67 / 0.54$ & $0.28 / 0.74 / 0.41$ & $0.45 / 0.20 / 0.27$ & $0.33 / 0.27 / 0.30$ & $0.24 / 0.34 / 0.28$ \\
 CoG+MonoT5 & $\textbf{0.72} / 0.54 / \textbf{0.62}$ & $\textbf{0.49} / 0.66 / \textbf{0.56}$ & $\textbf{0.30} / 0.75 / \textbf{0.43}$ & $0.46 / 0.20 / 0.28$ & $\textbf{0.35} / \textbf{0.28} / \textbf{0.31}$ & $\textbf{0.25} / \textbf{0.35} / \textbf{0.29}$ \\
\bottomrule
\end{tabular}
}
    \caption{Retrieval based attributor results}
    \label{tab:retriever}
\end{table*}

\begin{table}[t]
    \centering
    \scriptsize
\begin{tabular}{c|c|c}
\toprule  & { Verifiability } & { QASPER } \\
\midrule Decomposer+Attributor & P/R/F1 & P/R/F1 \\
\midrule  NIL+GPT 4 & 0.26/0.76/0.39 & 0.18/0.15/0.16 \\
 FActScore+GPT 4 & 0.18/0.69/0.29 & 0.17/0.27/0.21 \\
 (CoG-neg.)+GPT 4 & 0.26/0.76/0.39 & 0.19/0.29/0.23 \\
  CoG+GPT 4 & \textbf{0.29}/\textbf{0.79}/\textbf{0.42} & \textbf{0.22}/\textbf{0.32}/\textbf{0.26} \\
\midrule NIL+GPT 3.5 & 0.25/0.75/0.37 & 0.16/0.15/0.16 \\
 FActScore+GPT 3.5 & 0.17/0.69/0.27 & 0.15/0.27/0.20 \\
 (CoG – neg.) + GPT 3.5 & 0.27/0.75/0.40 & 0.18/0.27/0.22 \\
 CoG+GPT 3.5 & \textbf{0.28}/\textbf{0.77}/\textbf{0.41} & \textbf{0.22}/\textbf{0.31}/\textbf{0.26} \\
\midrule  NIL+LLaMa 2 (70 B) & 0.20/0.73/0.32 & 0.13/0.10/0.11 \\
  FActScore+LLaMa 2 (70 B) & 0.15/0.66/0.25 & 0.12/0.25/0.16 \\
  (CoG-neg.) + LLaMa 2 (70 B) & 0.22/0.75/0.34 & 0.17/0.26/0.21 \\
  CoG+LLaMa 2 (70 B) & \textbf{0.23}/\textbf{0.77}/\textbf{0.36} & \textbf{0.20}/\textbf{0.28}/\textbf{0.23} \\
\midrule NIL+LLaMa 2 (13 B) & 0.19/0.72/0.30 & 0.11/0.09/0.10 \\
 FActScore+LLaMa 2 (13 B) & 0.13/0.63/0.22 & 0.10/0.24/0.14 \\
 (CoG-neg.) + LLaMa 2 (13 B) & 0.21/0.73/0.33 & 0.16/0.25/0.20 \\
 CoG+LLaMa 2 (13 B) & \textbf{0.22}/\textbf{0.74}/\textbf{0.34} & \textbf{0.18}/\textbf{0.25}/\textbf{0.21} \\
\bottomrule
\end{tabular}
    \caption{LLM based attribution results}
    \label{tab:llm}
\end{table}
\section{Experiments and Evaluation}

\subsection{Baselines and Evaluation Strategy}

We evaluate the performance of three retrieval-based systems and four large language models (LLMs) as attributors. The retrieval methods include: BM25, a sparse model; GTR, a dense model \cite{ni-etal-2022-large}; and MonoT5 \cite{pradeep2021expandomonoduo}. We explore the capabilities of four LLMs: GPT-4 (\texttt{gpt-4}), GPT-3.5 (\texttt{gpt-3.5-turbo}), LLaMa 2 70B (\texttt{llama-2-70b-chat-hf}), and LLaMa 2 13B (\texttt{llama-2-13b-chat-hf}) \cite{touvron2023llama}. We use GPT-4 (\texttt{gpt-4}) for obtaining decomposition. We use NVIDIA A100 GPU to run inference for LLaMa 2 models.
We do not provide retrieve-and-read-based baselines ~\citep{guu2020realm,borgeaud2022improving,izacard2022few} as they generate answers along with attributions, whereas our task assumes answer as an input.
Table \ref{tab:retriever} and Table \ref{tab:llm} tabulates the results for retrieval based and LLM based attributors respectively. NIL refers to using an answer sentence as an information unit, FActScore refers to using fine grained decomposition as information units and CoG refers to question contextualised coarse grain decomposition as information units. For all the LLM based attributions, we take the top 100 sentences retrieved from BM25 to fit within the context limit. We keep the retrieved evidences same across all settings for a fair evaluation.

\subsection{Evaluation Measures}
For the retriever based attributors, since we get a score for a query and evidence, we report precision (P), recall (R), and F1 scores of top 1, 2, and 4 predicted attributions per sentence of an answer. For the LLM based attributors, we ask the LLM to output only highly relevant evidences. Since there is no score between the query (information unit) and evidence, we collect all the shortlisted evidences and report the precision, recall and F1 scores. To ensure a reliable evaluation, we do not consider the samples where answer sentences are an exact match from from the documents. 
For Verifiability dataset, we report attributions on sentence level for each sentence in the answer and for QASPER we report attributions on sentence level for the whole answer.

\section{Results and Discussion}
\begin{table*}
    \centering
    \scalebox{0.58}{
\begin{tabular}{c|c|c|c|c|c}
\hline \multirow{3}{*}{\begin{tabular}{l} 
Question \\
Answer \\ \\
GT Attribution \\
\end{tabular}} & \multicolumn{5}{|c}{ "paint cast iron" } \\
\hline & \multicolumn{5}{|c}{\begin{tabular}{l} 
"To paint cast iron, you should first coat it with oil-based primer to create a smooth surface \\ and help the paint adhere. You can find cast iron paint on Amazon."
\end{tabular}} \\
\hline & CoG Decompositions & NIL-MonoT5 & CoG-MonoT5 & NIL-GPT4 & CoG-GPT4 \\
\hline \begin{tabular}{l} 
["Coat the cast \\ 
iron with oil- \\ 
based primer.', \\
"Priming the metal \\
creates a smooth \\
surface and will \\ 
help the paint adhere."]
\end{tabular} & \begin{tabular}{l} 
["To paint cast \\ 
iron, you should \\ 
first coat it with \\
oil-based primer.", \\ 
"The oil-based \\ 
primer helps create \\
a smooth surface \\
and help the paint \\
adhere."]
\end{tabular} & \begin{tabular}{l} 
["If you're working \\
with a smaller \\ 
piece of cast \\
iron, you can \\ 
wipe it down \\
with a damp rag, \\
instead.", \\
"Apply oil-based \\
paint to the \\
cast iron."]
\end{tabular} & \begin{tabular}{l} 
["Coat the cast \\
iron with oil- \\
based primer.",\\
"If you're working \\
with a smaller \\
piece of cast \\
iron, you can \\
wipe it down \\
with a damp \\
rag, instead."]
\end{tabular} & \begin{tabular}{l}
["Priming the metal \\
creates a smooth \\
surface and will \\
help the paint \\
adhere.",\\
"Apply oil-based \\
paint to the \\
cast iron.", \\
"Coat the cast \\
iron with oil\\
-based primer."]
\end{tabular} & \begin{tabular}{l} 
["Coat the cast iron \\
with oil-based \\
primer.", \\
"Priming the metal \\
creates a smooth \\
surface and will \\
help the paint \\
adhere."]
\end{tabular} \\
\hline [] & \begin{tabular}{l} 
["You can find \\
cast iron paint \\ 
on Amazon."]
\end{tabular} & \begin{tabular}{l} 
["Read on for \\
our complete guide \\
to painting cast \\
iron easily at home."]
\end{tabular} & \begin{tabular}{l} 
["Read on for \\
our complete guide \\
to painting cast \\
iron easily at home."]
\end{tabular} & [] & [] \\
\hline
\end{tabular}
}
\caption{Qualitative example of how decomposition affects retrieval based attributor and LLM based attributor. GT refers to ground truth. Each row depicts an answer part and respective decompositions and attributions for each method.}
    \label{tab:ret_llm}
\end{table*}

In Table \ref{tab:retriever}, while using CoG as the decomposer for retrieval based methods, we observe increased precision and overall F1 score on the Citation Verifiability dataset. The observed increase in precision on the Verifiability dataset is because CoG decomposer does not provide information units for the sentences that do not require attribution. 

We notice comparable results over the QASPER dataset. This dataset majorly contains extractive and short sentences. The average length of extractive answers in the dataset is 14.4 and that of abstractive answers in 15.6 words. Due to the short and highly extractive nature of the dataset, it is likely that using answer sentences as information unit suffices. 

Interestingly, we observe that on QASPER, GTR and MonoT5 have comparable performance while using CoG and single answer sentence as an information unit. Whereas, in BM25 there is a slight reduction in scores. This is likely due to the way the answer is decomposed or \textit{augmented} when the question is taken into context. This impacts retrieval methods like BM25 where the frequency of words are taken into account to calculate the score. This motivates the use of embedding based retrievers as attributors over term frequency based retrievers. 

In Table \ref{tab:llm}, when utilizing CoG as the decomposer, we observe improvements across all LLMs. \cite{asher2023limits} demonstrated that LLMs operate without formal guarantees for tasks requiring entailments and in-depth language comprehension. Prompting techniques such as Chain-of-Thought (CoT) \cite{wei2023chainofthought} are well-known to enhance the performance of LLMs. The improved performance with the use of CoG proves that providing decomposed answers in the form of information units likely simplifies the task of finding evidence that entails the specified information unit. 

FActScore based decomposition performs poorly across all settings. The fine grain decompositions do not capture the information that is required to be attributed. Table \ref{tab:decomp_qualitative} shows the large number of decompositions obtained for an answer. Intuitively, having a large number of decompositions should result in higher recall and lower precision. Yet, this trend is not observed. This indicates that the information that the answer part conveys gets diluted when broken down into finer parts. This impacts the retriever and LLM based attributors to perform well. We also observe that CoG without negative sampling performs slightly poor than with negative sampling in all cases.

We observe that identifying relevant information helps MonoT5 make better attributions compared to using answer sentences as queries. Retrieval-based attributors rely on query and evidence scores, which can lead to irrelevant matches, especially since finding an appropriate threshold is challenging and dataset-specific. Unlike retrieval methods, LLMs consider question context, resulting in more accurate attributions. However, it is interesting to observe that when the information unit is given as a complete sentence, LLM struggles to find the precise set of evidences. Examples for these observations are present in Table \ref{tab:ret_llm} (Appendix \ref{sec:ret_llm}).

\begin{table*}
    \centering
    \scalebox{0.65}{
    \begin{tabular}{c|p{19.5cm}}
        \toprule
         Question & paint cast iron \\
         \midrule
         Answer & \textcolor{red}{Are you looking for information on how to paint cast iron? If so, I found a helpful article on wikiHow that provides a step-by-step guide on how to paint cast iron.} The process involves prepping the cast iron by cleaning it and removing any rust or old paint. Then, you can prime and paint the cast iron using an oil-based primer and paint. \textcolor{red}{Would you like more information on this topic?} \\
         \midrule
         FActScore & [`The person is looking for information.', `The person is looking for information on how to paint cast iron.'], [`There is a helpful article on wikiHow.',`The article provides a step-by-step guide.', `The article is about how to paint cast iron.'], [`The process involves prepping cast iron.', `Prepping cast iron involves cleaning it.', `Prepping cast iron involves removing rust.', `Prepping cast iron involves removing old paint.'], [`You can prime with an oil-based primer.', `You can paint with an oil-based paint.', `You can prime and paint a cast iron.'], [`This is a question.', `The topic referenced is unclear.]\\
         \midrule
         CoG - Question & [], [`There is a helpful article on wikiHow that provides a step-by-step guide on how to paint cast iron.'], [`The process involves prepping the cast iron by cleaning it', `The process involves prepping the cast iron by removing any rust or old paint.'], [`Prime the cast iron using an oil-based primer.', `Paint the cast iron using an oil-based paint'], [] \\
         \midrule
         CoG & [], [], [`The process involves prepping the cast iron by cleaning it', `The process involves prepping the cast iron by removing any rust or old paint.'], [`Prime the cast iron using an oil-based primer.', `Paint the cast iron using an oil-based paint'], [] \\
         \bottomrule
    \end{tabular}
    }
    \caption{Example from Citation Verifiability dataset: In the answer, portions highlighted in red do not need attributions. Lists show the decomposition outputs for each answer part. CoG - Question denotes coarse-grain decomposition without the question in context.}
    \label{tab:decomp_qualitative}
\end{table*}

\begin{table}[t]
    \centering
    \scalebox{0.75}{
    \begin{tabular}{p{5cm}|p{2cm}|p{2cm}}
    \toprule
      Evidence & NIL+MonoT5 & CoG+MonoT5 \\
    \midrule
    "WNUT16: WNUT16 was a shared task on Named Entity Recognition over Twitter BIBREF10." & -0.043 & -0.021\\
    \bottomrule
    \end{tabular}
    }
    \caption{Example of retriever score getting affected while using answer part as \textit{iu} vs using decomposed \textit{iu}.}
    \label{tab:dataset}
\end{table}

\subsection{Ablation Study}

\subsubsection{Analysis of Decomposers}
\begin{figure}[t!]
    \centering
    \includegraphics[width=0.8\columnwidth]{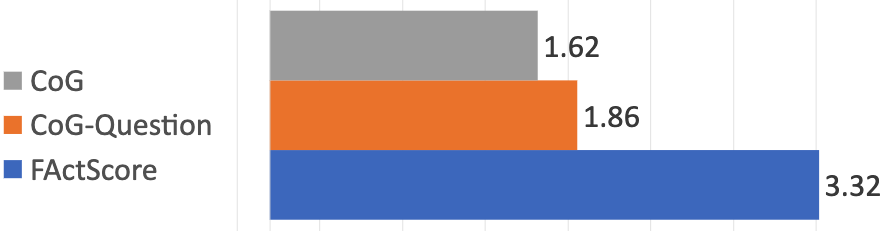}
    \caption{Average number of decomposition per sentence using each method.}
    \label{fig:per_sentence}
\end{figure}

In the Citation Verifiability dataset, ground truth citations are available only for answer parts that have information relevant to the question. We assess sentence attributions using FActScore, CoG-Question (coarse grain decomposition without the question as context), and CoG as decomposers. In Verifiablility datset, \textbf{573} sentences \textbf{do not} require attributions. Using FActScore, we attribute \textbf{509} of them compared to \textbf{491} using CoG-Question and \textbf{473} using CoG decomposition. Table \ref{tab:decomp_qualitative} provides a qualitative example decomposition obtained. We observe that fine-grained attribution may lead to information duplication across multiple units, resulting in a higher number of decomposition per sentence, as shown in Figure \ref{fig:per_sentence}.

\paragraph{Human evaluation.} We conduct a human survey to validate the quality of decomposition. The objective is to understand what specific facts within an answer are essential and should be credited as factual references in a document. We ask 3 annotators of similar backgrounds (Indian origin, above undergraduate studies, fluent in English). They are provided with 120 examples each, along with question, answer, FActScore and CoG decomposition. We provide the instruction set that is given to the LLM, so humans can validate whether the decomposition adhere to the requirements. 

In terms of alignment of answer decomposition for the task of attribution, our outputs are marked better than the baseline in 80\% cases. The inter-annotator agreement \cite{Krippendorff1970} agreement for is 0.68, indicating a strong agreement among annotators. The details of the survey are provided in Appendix \ref{sec:survey}.

\paragraph{Classifier.} We check the number of answer sentences that are returned without decomposition on the dev set of QASPER and Verifiability dataset. The classifier correctly identifies 25 out of 158 answer sentences (17\%) on QASPER and 22 out of 184 answer sentences in Verifiability (12\%) as sentences that do not require any decomposition. The classifier does not classify any answer sentence that undergoes decomposition as \textit{a simple sentence}.

\subsection{Effect of Decomposition on Retrievers}
Complex answer sentences in retrieval systems can impact the scores obtained for evidence sentences. For instance, in Table \ref{tab:dataset}, the answer comprising multiple facts mapped to different evidence may result in lower scores for the correct evidence compared to using decomposed information units as search queries.

\subsubsection{Effect of Decomposition on LLMs}
Table \ref{tab:llm} shows improved performance when CoG decomposition are used to guide the LLM. We observe that while attributions obtained for a complex answer (without decomposition) aligns with the answer on a high level, it does not provide concrete support to the information present in the answer. Whereas, in CoG+GPT4, we notice a more nuanced evidence retrieval because the task of searching information is broken down into smaller pieces. Examples for these observation are present in Table \ref{tab:llm_decomp_abl} (Appendix \ref{sec:abl}).

\section{Conclusion}

In this paper, we raise research questions regarding the consideration of post-hoc attributions as a grounding task rather than a naive mapping task. On this basis, we propose and explore a novel approach to the factual decomposition of generated answers for attribution, employing template-based instructional tuning. We empirically establish the fact that our proposed granular-level Coarse Grained Decomposition (CoG) helps identify the spans of answers that need decomposition by following the semantic context inferred from the question asked. We also qualitatively and empirically establish that using Language Models (LLMs) as attributors provides the breathing space to consider attribution as contextual semantic grounding rather than performing as a retrieval and mapping task. Through various ablation studies, we establish that on an extractive dataset, retriever-based algorithms can perform better by incorporating our Coarse Grained Decomposition as input.
\section{Limitations}
We present promising results in enhancing post-hoc attributions for comprehending lengthy documents. Adopting a post-hoc perspective provides research opportunities when dealing with outputs from opaque models. Currently, our focus is solely on post-hoc attributions through a unimodal lens, involving textual input and textual output. The transition to multimodal attributions would present additional challenges. An intriguing avenue for research involves exploring attributions from tables, charts, and images and performing reasoning over them.

In our post-hoc approach, we identify only those parts with supporting evidence but do not address how to mitigate unsupported claims. This creates opportunities to explore the incorporation of feedback loops for unattributable answer parts, aiming to generate more reliable answers.

\bibliography{bibs/anthology,bibs/custom}

\begin{thebibliography}{41}
\expandafter\ifx\csname natexlab\endcsname\relax\def\natexlab#1{#1}\fi

\bibitem[{Bosselut et~al.(2019)Bosselut, Rashkin, Sap, Malaviya, Celikyilmaz,
  and Choi}]{bosselut-etal-2019-comet}
Antoine Bosselut, Hannah Rashkin, Maarten Sap, Chaitanya Malaviya, Asli
  Celikyilmaz, and Yejin Choi. 2019.
\newblock \href {https://doi.org/10.18653/v1/P19-1470} {{COMET}: Commonsense
  transformers for automatic knowledge graph construction}.
\newblock In \emph{Proceedings of the 57th Annual Meeting of the Association
  for Computational Linguistics}, pages 4762--4779, Florence, Italy.
  Association for Computational Linguistics.

\bibitem[{Bouraoui et~al.(2020)Bouraoui, Camacho-Collados, and
  Schockaert}]{bouraoui2020inducing}
Zied Bouraoui, Jose Camacho-Collados, and Steven Schockaert. 2020.
\newblock Inducing relational knowledge from {BERT}.
\newblock In \emph{Proceedings of the AAAI Conference on Artificial
  Intelligence}, volume~34, pages 7456--7463.

\bibitem[{Chambers(2017)}]{chambers2017behind}
Nathanael Chambers. 2017.
\newblock Behind the scenes of an evolving event cloze test.
\newblock In \emph{Proceedings of the 2nd Workshop on Linking Models of
  Lexical, Sentential and Discourse-level Semantics}, pages 41--45.

\bibitem[{Chambers and Jurafsky(2008)}]{chambers2008unsupervised}
Nathanael Chambers and Dan Jurafsky. 2008.
\newblock Unsupervised learning of narrative event chains.
\newblock In \emph{Proceedings of ACL-08: HLT}, pages 789--797.

\bibitem[{Chambers and Jurafsky(2009)}]{chambers2009unsupervised}
Nathanael Chambers and Dan Jurafsky. 2009.
\newblock Unsupervised learning of narrative schemas and their participants.
\newblock In \emph{Proceedings of the Joint Conference of the 47th Annual
  Meeting of the ACL and the 4th International Joint Conference on Natural
  Language Processing of the AFNLP}, pages 602--610.

\bibitem[{Cohen(1960)}]{cohen1960coefficient}
Jacob Cohen. 1960.
\newblock A coefficient of agreement for nominal scales.
\newblock \emph{Educational and psychological measurement}, 20(1):37--46.

\bibitem[{Devlin et~al.(2018)Devlin, Chang, Lee, and
  Toutanova}]{devlin2018bert}
Jacob Devlin, Ming-Wei Chang, Kenton Lee, and Kristina Toutanova. 2018.
\newblock {BERT}: Pre-training of deep bidirectional transformers for language
  understanding.
\newblock \emph{arXiv preprint arXiv:1810.04805}.

\bibitem[{Dolan and Brockett(2005)}]{dolan2005automatically}
William~B Dolan and Chris Brockett. 2005.
\newblock Automatically constructing a corpus of sentential paraphrases.
\newblock In \emph{Proceedings of the Third International Workshop on
  Paraphrasing (IWP2005)}.

\bibitem[{Feldman et~al.(2019)Feldman, Davison, and
  Rush}]{feldman2019commonsense}
Joshua Feldman, Joe Davison, and Alexander~M Rush. 2019.
\newblock Commonsense knowledge mining from pretrained models.
\newblock \emph{arXiv preprint arXiv:1909.00505}.

\bibitem[{Gordon and Van~Durme(2013)}]{gordon2013reporting}
Jonathan Gordon and Benjamin Van~Durme. 2013.
\newblock Reporting bias and knowledge acquisition.
\newblock In \emph{Proceedings of the 2013 workshop on Automated knowledge base
  construction}, pages 25--30.

\bibitem[{Holtzman et~al.(2019)Holtzman, Buys, Du, Forbes, and
  Choi}]{holtzman2019curious}
Ari Holtzman, Jan Buys, Li~Du, Maxwell Forbes, and Yejin Choi. 2019.
\newblock The curious case of neural text degeneration.
\newblock \emph{arXiv preprint arXiv:1904.09751}.

\bibitem[{Lin et~al.(2020)Lin, Lee, Khanna, and Ren}]{lin2020birds}
Bill~Yuchen Lin, Seyeon Lee, Rahul Khanna, and Xiang Ren. 2020.
\newblock Birds have four legs?! numersense: Probing numerical commonsense
  knowledge of pre-trained language models.
\newblock \emph{arXiv preprint arXiv:2005.00683}.

\bibitem[{Liu et~al.(2019)Liu, Ott, Goyal, Du, Joshi, Chen, Levy, Lewis,
  Zettlemoyer, and Stoyanov}]{liu2019roberta}
Yinhan Liu, Myle Ott, Naman Goyal, Jingfei Du, Mandar Joshi, Danqi Chen, Omer
  Levy, Mike Lewis, Luke Zettlemoyer, and Veselin Stoyanov. 2019.
\newblock Ro{BERT}a: A {R}obustly {O}ptimized {BERT} {P}retraining {A}pproach.
\newblock \emph{arXiv preprint arXiv:1907.11692}.

\bibitem[{Lyu et~al.(2020)Lyu, Zhang, and
  Callison-Burch}]{lyu-zhang-wikihow:2020}
Qing Lyu, Li~Zhang, and Chris Callison-Burch. 2020.
\newblock \href
  {http://www.cis.upenn.edu/~ccb/publications/reasoning-about-goals-with-wikihow.pdf}
  {Reasoning about goals, steps, and temporal ordering with wikihow}.
\newblock In \emph{Proceedings of The 2020 Conference on Empirical Methods In
  Natural Language Proceedings (EMNLP)}.

\bibitem[{Miikkulainen(1995)}]{miikkulainen1995script}
Risto Miikkulainen. 1995.
\newblock Script-based inference and memory retrieval in subsymbolic story
  processing.
\newblock \emph{Applied Intelligence}, 5(2):137--163.

\bibitem[{Modi et~al.(2017)Modi, Anikina, Ostermann, and
  Pinkal}]{modi2017inscript}
Ashutosh Modi, Tatjana Anikina, Simon Ostermann, and Manfred Pinkal. 2017.
\newblock Inscript: Narrative texts annotated with script information.
\newblock \emph{arXiv preprint arXiv:1703.05260}.

\bibitem[{Modi and Titov(2014)}]{modi2014inducing}
Ashutosh Modi and Ivan Titov. 2014.
\newblock Inducing neural models of script knowledge.
\newblock In \emph{Proceedings of the Eighteenth Conference on Computational
  Natural Language Learning}, pages 49--57.

\bibitem[{Mostafazadeh et~al.(2016)Mostafazadeh, Chambers, He, Parikh, Batra,
  Vanderwende, Kohli, and Allen}]{mostafazadeh2016corpus}
Nasrin Mostafazadeh, Nathanael Chambers, Xiaodong He, Devi Parikh, Dhruv Batra,
  Lucy Vanderwende, Pushmeet Kohli, and James Allen. 2016.
\newblock A corpus and cloze evaluation for deeper understanding of commonsense
  stories.
\newblock In \emph{Proceedings of the 2016 Conference of the North American
  Chapter of the Association for Computational Linguistics: Human Language
  Technologies}, pages 839--849.

\bibitem[{Mueller(2004)}]{mueller2004understanding}
Erik~T Mueller. 2004.
\newblock Understanding script-based stories using commonsense reasoning.
\newblock \emph{Cognitive Systems Research}, 5(4):307--340.

\bibitem[{Ostermann(2020)}]{ostermann2020script}
Simon Ostermann. 2020.
\newblock Script knowledge for natural language understanding.

\bibitem[{Ostermann et~al.(2018)Ostermann, Modi, Roth, Thater, and
  Pinkal}]{ostermann2018mcscript}
Simon Ostermann, Ashutosh Modi, Michael Roth, Stefan Thater, and Manfred
  Pinkal. 2018.
\newblock Mcscript: A novel dataset for assessing machine comprehension using
  script knowledge.
\newblock \emph{arXiv preprint arXiv:1803.05223}.

\bibitem[{Ostermann et~al.(2019)Ostermann, Roth, and
  Pinkal}]{ostermann2019mcscript2}
Simon Ostermann, Michael Roth, and Manfred Pinkal. 2019.
\newblock Mcscript2. 0: A machine comprehension corpus focused on script events
  and participants.
\newblock \emph{arXiv preprint arXiv:1905.09531}.

\bibitem[{Papineni et~al.(2002)Papineni, Roukos, Ward, and
  Zhu}]{papineni-etal-2002-bleu}
Kishore Papineni, Salim Roukos, Todd Ward, and Wei-Jing Zhu. 2002.
\newblock \href {https://doi.org/10.3115/1073083.1073135} {{B}leu: a method for
  automatic evaluation of machine translation}.
\newblock In \emph{Proceedings of the 40th Annual Meeting of the Association
  for Computational Linguistics}, pages 311--318, Philadelphia, Pennsylvania,
  USA. Association for Computational Linguistics.

\bibitem[{Petroni et~al.(2020)Petroni, Lewis, Piktus, Rockt{\"a}schel, Wu,
  Miller, and Riedel}]{petroni2020context}
Fabio Petroni, Patrick Lewis, Aleksandra Piktus, Tim Rockt{\"a}schel, Yuxiang
  Wu, Alexander~H Miller, and Sebastian Riedel. 2020.
\newblock How context affects language models' factual predictions.
\newblock \emph{arXiv preprint arXiv:2005.04611}.

\bibitem[{Pichotta and Mooney(2016)}]{pichotta2016using}
Karl Pichotta and Raymond~J Mooney. 2016.
\newblock Using sentence-level lstm language models for script inference.
\newblock \emph{arXiv preprint arXiv:1604.02993}.

\bibitem[{Radford et~al.(2019)Radford, Wu, Child, Luan, Amodei, and
  Sutskever}]{radford2019language}
Alec Radford, Jeffrey Wu, Rewon Child, David Luan, Dario Amodei, and Ilya
  Sutskever. 2019.
\newblock Language models are unsupervised multitask learners.
\newblock \emph{OpenAI blog}, 1(8):9.

\bibitem[{Regneri et~al.(2010)Regneri, Koller, and
  Pinkal}]{regneri-etal-2010-learning}
Michaela Regneri, Alexander Koller, and Manfred Pinkal. 2010.
\newblock \href {https://www.aclweb.org/anthology/P10-1100} {Learning script
  knowledge with web experiments}.
\newblock In \emph{Proceedings of the 48th Annual Meeting of the Association
  for Computational Linguistics}, pages 979--988, Uppsala, Sweden. Association
  for Computational Linguistics.

\bibitem[{Rudinger et~al.(2015)Rudinger, Rastogi, Ferraro, and
  Van~Durme}]{rudinger-etal-2015-script}
Rachel Rudinger, Pushpendre Rastogi, Francis Ferraro, and Benjamin Van~Durme.
  2015.
\newblock \href {https://doi.org/10.18653/v1/D15-1195} {Script induction as
  language modeling}.
\newblock In \emph{Proceedings of the 2015 Conference on Empirical Methods in
  Natural Language Processing}, pages 1681--1686, Lisbon, Portugal. Association
  for Computational Linguistics.

\bibitem[{Sakaguchi et~al.(2021)Sakaguchi, Bhagavatula, Bras, Tandon, Clark,
  and Choi}]{sakaguchi2021proscript}
Keisuke Sakaguchi, Chandra Bhagavatula, Ronan~Le Bras, Niket Tandon, Peter
  Clark, and Yejin Choi. 2021.
\newblock proscript: Partially ordered scripts generation via pre-trained
  language models.
\newblock \emph{arXiv preprint arXiv:2104.08251}.

\bibitem[{Sap et~al.(2019)Sap, Le~Bras, Allaway, Bhagavatula, Lourie, Rashkin,
  Roof, Smith, and Choi}]{sap2019atomic}
Maarten Sap, Ronan Le~Bras, Emily Allaway, Chandra Bhagavatula, Nicholas
  Lourie, Hannah Rashkin, Brendan Roof, Noah~A Smith, and Yejin Choi. 2019.
\newblock Atomic: An atlas of machine commonsense for if-then reasoning.
\newblock In \emph{Proceedings of the AAAI Conference on Artificial
  Intelligence}, volume~33, pages 3027--3035.

\bibitem[{Schank and Abelson(1975)}]{schank1975scripts}
Roger~C Schank and Robert~P Abelson. 1975.
\newblock Scripts, plans, and knowledge.
\newblock In \emph{IJCAI}, volume~75, pages 151--157.

\bibitem[{Shwartz et~al.(2020)Shwartz, West, Bras, Bhagavatula, and
  Choi}]{shwartz2020unsupervised}
Vered Shwartz, Peter West, Ronan~Le Bras, Chandra Bhagavatula, and Yejin Choi.
  2020.
\newblock Unsupervised commonsense question answering with self-talk.
\newblock \emph{arXiv preprint arXiv:2004.05483}.

\bibitem[{Singh et~al.(2002)Singh, Lin, Mueller, Lim, Perkins, and
  Zhu}]{singh2002open}
Push Singh, Thomas Lin, Erik~T Mueller, Grace Lim, Travell Perkins, and Wan~Li
  Zhu. 2002.
\newblock Open mind common sense: Knowledge acquisition from the general
  public.
\newblock In \emph{OTM Confederated International Conferences" On the Move to
  Meaningful Internet Systems"}, pages 1223--1237. Springer.

\bibitem[{Wanzare et~al.(2017{\natexlab{a}})Wanzare, Zarcone, Thater, and
  Pinkal}]{wanzare2017inducing}
Lilian Wanzare, Alessandra Zarcone, Stefan Thater, and Manfred Pinkal.
  2017{\natexlab{a}}.
\newblock Inducing script structure from crowdsourced event descriptions via
  semi-supervised clustering.

\bibitem[{Wanzare et~al.(2017{\natexlab{b}})Wanzare, Zarcone, Thater, and
  Pinkal}]{wanzare-etal-2017-inducing}
Lilian Wanzare, Alessandra Zarcone, Stefan Thater, and Manfred Pinkal.
  2017{\natexlab{b}}.
\newblock \href {https://doi.org/10.18653/v1/W17-0901} {Inducing script
  structure from crowdsourced event descriptions via semi-supervised
  clustering}.
\newblock In \emph{Proceedings of the 2nd Workshop on Linking Models of
  Lexical, Sentential and Discourse-level Semantics}, pages 1--11, Valencia,
  Spain. Association for Computational Linguistics.

\bibitem[{Wanzare et~al.(2016)Wanzare, Zarcone, Thater, and
  Pinkal}]{wanzare2016crowdsourced}
Lilian~DA Wanzare, Alessandra Zarcone, Stefan Thater, and Manfred Pinkal. 2016.
\newblock A crowdsourced database of event sequence descriptions for the
  acquisition of high-quality script knowledge.

\bibitem[{Weir et~al.(2020)Weir, Poliak, and Van~Durme}]{weir2020probing}
Nathaniel Weir, Adam Poliak, and Benjamin Van~Durme. 2020.
\newblock Probing neural language models for human tacit assumptions.
\newblock CogSci.

\bibitem[{Wolf et~al.(2020)Wolf, Debut, Sanh, Chaumond, Delangue, Moi, Cistac,
  Rault, Louf, Funtowicz, Davison, Shleifer, von Platen, Ma, Jernite, Plu, Xu,
  Le~Scao, Gugger, Drame, Lhoest, and Rush}]{wolf-etal-2020-transformers}
Thomas Wolf, Lysandre Debut, Victor Sanh, Julien Chaumond, Clement Delangue,
  Anthony Moi, Pierric Cistac, Tim Rault, Remi Louf, Morgan Funtowicz, Joe
  Davison, Sam Shleifer, Patrick von Platen, Clara Ma, Yacine Jernite, Julien
  Plu, Canwen Xu, Teven Le~Scao, Sylvain Gugger, Mariama Drame, Quentin Lhoest,
  and Alexander Rush. 2020.
\newblock \href {https://doi.org/10.18653/v1/2020.emnlp-demos.6} {Transformers:
  State-of-the-art natural language processing}.
\newblock In \emph{Proceedings of the 2020 Conference on Empirical Methods in
  Natural Language Processing: System Demonstrations}, pages 38--45, Online.
  Association for Computational Linguistics.

\bibitem[{Zhang et~al.(2020)Zhang, Chen, Wang, Song, and
  Roth}]{zhang2020analogous}
Hongming Zhang, Muhao Chen, Haoyu Wang, Yangqiu Song, and Dan Roth. 2020.
\newblock Analogous process structure induction for sub-event sequence
  prediction.
\newblock \emph{arXiv preprint arXiv:2010.08525}.

\bibitem[{Zhou et~al.(2020)Zhou, Ning, Khashabi, and Roth}]{zhou2020temporal}
Ben Zhou, Qiang Ning, Daniel Khashabi, and Dan Roth. 2020.
\newblock Temporal common sense acquisition with minimal supervision.
\newblock \emph{arXiv preprint arXiv:2005.04304}.

\bibitem[{Zhou et~al.(2019)Zhou, Shah, and Schockaert}]{zhou2019learning}
Yilun Zhou, Julie~A Shah, and Steven Schockaert. 2019.
\newblock Learning household task knowledge from wikihow descriptions.
\newblock \emph{arXiv preprint arXiv:1909.06414}.

\end{thebibliography}


\begin{thebibliography}{41}
\expandafter\ifx\csname natexlab\endcsname\relax\def\natexlab#1{#1}\fi

\bibitem[{Ancona et~al.(2017)Ancona, Ceolini, {\"{O}}ztireli, and Gross}]{DBLP:journals/corr/abs-1711-06104}
Marco Ancona, Enea Ceolini, A.~Cengiz {\"{O}}ztireli, and Markus~H. Gross. 2017.
\newblock \href {http://arxiv.org/abs/1711.06104} {A unified view of gradient-based attribution methods for deep neural networks}.
\newblock \emph{CoRR}, abs/1711.06104.

\bibitem[{Asher et~al.(2023)Asher, Bhar, Chaturvedi, Hunter, and Paul}]{asher2023limits}
Nicholas Asher, Swarnadeep Bhar, Akshay Chaturvedi, Julie Hunter, and Soumya Paul. 2023.
\newblock \href {http://arxiv.org/abs/2306.12213} {Limits for learning with language models}.

\bibitem[{Bohnet et~al.(2023)Bohnet, Tran, Verga, Aharoni, Andor, Soares, Ciaramita, Eisenstein, Ganchev, Herzig, Hui, Kwiatkowski, Ma, Ni, Saralegui, Schuster, Cohen, Collins, Das, Metzler, Petrov, and Webster}]{bohnet2023attributed}
Bernd Bohnet, Vinh~Q. Tran, Pat Verga, Roee Aharoni, Daniel Andor, Livio~Baldini Soares, Massimiliano Ciaramita, Jacob Eisenstein, Kuzman Ganchev, Jonathan Herzig, Kai Hui, Tom Kwiatkowski, Ji~Ma, Jianmo Ni, Lierni~Sestorain Saralegui, Tal Schuster, William~W. Cohen, Michael Collins, Dipanjan Das, Donald Metzler, Slav Petrov, and Kellie Webster. 2023.
\newblock \href {http://arxiv.org/abs/2212.08037} {Attributed question answering: Evaluation and modeling for attributed large language models}.

\bibitem[{Borgeaud et~al.(2022)Borgeaud, Mensch, Hoffmann, Cai, Rutherford, Millican, van~den Driessche, Lespiau, Damoc, Clark, de~Las~Casas, Guy, Menick, Ring, Hennigan, Huang, Maggiore, Jones, Cassirer, Brock, Paganini, Irving, Vinyals, Osindero, Simonyan, Rae, Elsen, and Sifre}]{borgeaud2022improving}
Sebastian Borgeaud, Arthur Mensch, Jordan Hoffmann, Trevor Cai, Eliza Rutherford, Katie Millican, George van~den Driessche, Jean-Baptiste Lespiau, Bogdan Damoc, Aidan Clark, Diego de~Las~Casas, Aurelia Guy, Jacob Menick, Roman Ring, Tom Hennigan, Saffron Huang, Loren Maggiore, Chris Jones, Albin Cassirer, Andy Brock, Michela Paganini, Geoffrey Irving, Oriol Vinyals, Simon Osindero, Karen Simonyan, Jack~W. Rae, Erich Elsen, and Laurent Sifre. 2022.
\newblock \href {http://arxiv.org/abs/2112.04426} {Improving language models by retrieving from trillions of tokens}.

\bibitem[{Dasigi et~al.(2021)Dasigi, Lo, Beltagy, Cohan, Smith, and Gardner}]{dasigi2021dataset}
Pradeep Dasigi, Kyle Lo, Iz~Beltagy, Arman Cohan, Noah~A. Smith, and Matt Gardner. 2021.
\newblock \href {http://arxiv.org/abs/2105.03011} {A dataset of information-seeking questions and answers anchored in research papers}.

\bibitem[{Dziri et~al.(2022)Dziri, Rashkin, Linzen, and Reitter}]{dziri-etal-2022-evaluating}
Nouha Dziri, Hannah Rashkin, Tal Linzen, and David Reitter. 2022.
\newblock \href {https://doi.org/10.1162/tacl_a_00506} {Evaluating attribution in dialogue systems: The {BEGIN} benchmark}.
\newblock \emph{Transactions of the Association for Computational Linguistics}, 10:1066--1083.

\bibitem[{Evans et~al.(2021)Evans, Cotton-Barratt, Finnveden, Bales, Balwit, Wills, Righetti, and Saunders}]{evans2021truthful}
Owain Evans, Owen Cotton-Barratt, Lukas Finnveden, Adam Bales, Avital Balwit, Peter Wills, Luca Righetti, and William Saunders. 2021.
\newblock \href {http://arxiv.org/abs/2110.06674} {Truthful ai: Developing and governing ai that does not lie}.

\bibitem[{Fan et~al.(2019)Fan, Jernite, Perez, Grangier, Weston, and Auli}]{fan-etal-2019-eli5}
Angela Fan, Yacine Jernite, Ethan Perez, David Grangier, Jason Weston, and Michael Auli. 2019.
\newblock \href {https://doi.org/10.18653/v1/P19-1346} {{ELI}5: Long form question answering}.
\newblock In \emph{Proceedings of the 57th Annual Meeting of the Association for Computational Linguistics}, pages 3558--3567, Florence, Italy. Association for Computational Linguistics.

\bibitem[{Funkquist et~al.(2023)Funkquist, Kuznetsov, Hou, and Gurevych}]{funkquist2023citebench}
Martin Funkquist, Ilia Kuznetsov, Yufang Hou, and Iryna Gurevych. 2023.
\newblock \href {http://arxiv.org/abs/2212.09577} {Citebench: A benchmark for scientific citation text generation}.

\bibitem[{Gao et~al.(2023{\natexlab{a}})Gao, Dai, Pasupat, Chen, Chaganty, Fan, Zhao, Lao, Lee, Juan et~al.}]{gao2023rarr}
Luyu Gao, Zhuyun Dai, Panupong Pasupat, Anthony Chen, Arun~Tejasvi Chaganty, Yicheng Fan, Vincent Zhao, Ni~Lao, Hongrae Lee, Da-Cheng Juan, et~al. 2023{\natexlab{a}}.
\newblock Rarr: Researching and revising what language models say, using language models.
\newblock In \emph{Proceedings of the 61st Annual Meeting of the Association for Computational Linguistics (Volume 1: Long Papers)}, pages 16477--16508.

\bibitem[{Gao et~al.(2023{\natexlab{b}})Gao, Yen, Yu, and Chen}]{gao2023enabling}
Tianyu Gao, Howard Yen, Jiatong Yu, and Danqi Chen. 2023{\natexlab{b}}.
\newblock \href {http://arxiv.org/abs/2305.14627} {Enabling large language models to generate text with citations}.

\bibitem[{Guo et~al.(2022)Guo, Schlichtkrull, and Vlachos}]{guo-etal-2022-survey}
Zhijiang Guo, Michael Schlichtkrull, and Andreas Vlachos. 2022.
\newblock \href {https://doi.org/10.1162/tacl_a_00454} {A survey on automated fact-checking}.
\newblock \emph{Transactions of the Association for Computational Linguistics}, 10:178--206.

\bibitem[{Guu et~al.(2020)Guu, Lee, Tung, Pasupat, and Chang}]{guu2020realm}
Kelvin Guu, Kenton Lee, Zora Tung, Panupong Pasupat, and Ming-Wei Chang. 2020.
\newblock \href {http://arxiv.org/abs/2002.08909} {Realm: Retrieval-augmented language model pre-training}.

\bibitem[{Holzinger et~al.(2021)Holzinger, Malle, Saranti, and Pfeifer}]{DBLP:journals/inffus/HolzingerMSP21}
Andreas Holzinger, Bernd Malle, Anna Saranti, and Bastian Pfeifer. 2021.
\newblock \href {https://doi.org/10.1016/J.INFFUS.2021.01.008} {Towards multi-modal causability with graph neural networks enabling information fusion for explainable {AI}}.
\newblock \emph{Inf. Fusion}, 71:28--37.

\bibitem[{Huang and Chang(2024)}]{huang-chang-2024-citation}
Jie Huang and Kevin Chang. 2024.
\newblock \href {https://doi.org/10.18653/v1/2024.findings-naacl.31} {Citation: A key to building responsible and accountable large language models}.
\newblock In \emph{Findings of the Association for Computational Linguistics: NAACL 2024}, pages 464--473, Mexico City, Mexico. Association for Computational Linguistics.

\bibitem[{Huang and Chang(2023)}]{huang2023citation}
Jie Huang and Kevin Chen-Chuan Chang. 2023.
\newblock \href {http://arxiv.org/abs/2307.02185} {Citation: A key to building responsible and accountable large language models}.

\bibitem[{Huang et~al.(2023)Huang, Yu, Ma, Zhong, Feng, Wang, Chen, Peng, Feng, Qin et~al.}]{huang2023survey}
Lei Huang, Weijiang Yu, Weitao Ma, Weihong Zhong, Zhangyin Feng, Haotian Wang, Qianglong Chen, Weihua Peng, Xiaocheng Feng, Bing Qin, et~al. 2023.
\newblock A survey on hallucination in large language models: Principles, taxonomy, challenges, and open questions.
\newblock \emph{arXiv preprint arXiv:2311.05232}.

\bibitem[{Izacard et~al.(2022)Izacard, Lewis, Lomeli, Hosseini, Petroni, Schick, Dwivedi-Yu, Joulin, Riedel, and Grave}]{izacard2022few}
Gautier Izacard, Patrick Lewis, Maria Lomeli, Lucas Hosseini, Fabio Petroni, Timo Schick, Jane Dwivedi-Yu, Armand Joulin, Sebastian Riedel, and Edouard Grave. 2022.
\newblock \href {http://arxiv.org/abs/2208.03299} {Atlas: Few-shot learning with retrieval augmented language models}.

\bibitem[{Krippendorff(1970)}]{Krippendorff1970}
Klaus Krippendorff. 1970.
\newblock Estimating the reliability, systematic error and random error of interval data.
\newblock \emph{Educational and Psychological Measurement}, 30(1):61--70.

\bibitem[{Kwiatkowski et~al.(2019)Kwiatkowski, Palomaki, Redfield, Collins, Parikh, Alberti, Epstein, Polosukhin, Devlin, Lee, Toutanova, Jones, Kelcey, Chang, Dai, Uszkoreit, Le, and Petrov}]{kwiatkowski-etal-2019-natural}
Tom Kwiatkowski, Jennimaria Palomaki, Olivia Redfield, Michael Collins, Ankur Parikh, Chris Alberti, Danielle Epstein, Illia Polosukhin, Jacob Devlin, Kenton Lee, Kristina Toutanova, Llion Jones, Matthew Kelcey, Ming-Wei Chang, Andrew~M. Dai, Jakob Uszkoreit, Quoc Le, and Slav Petrov. 2019.
\newblock \href {https://doi.org/10.1162/tacl_a_00276} {Natural questions: A benchmark for question answering research}.
\newblock \emph{Transactions of the Association for Computational Linguistics}, 7:452--466.

\bibitem[{Li et~al.(2022)Li, Ma, and Lin}]{li-etal-2022-encoder}
Minghan Li, Xueguang Ma, and Jimmy Lin. 2022.
\newblock \href {https://doi.org/10.18653/v1/2022.trustnlp-1.1} {An encoder attribution analysis for dense passage retriever in open-domain question answering}.
\newblock In \emph{Proceedings of the 2nd Workshop on Trustworthy Natural Language Processing (TrustNLP 2022)}, pages 1--11, Seattle, U.S.A. Association for Computational Linguistics.

\bibitem[{Litschko et~al.(2023)Litschko, Müller-Eberstein, van~der Goot, Weber, and Plank}]{litschko2023establishing}
Robert Litschko, Max Müller-Eberstein, Rob van~der Goot, Leon Weber, and Barbara Plank. 2023.
\newblock \href {http://arxiv.org/abs/2310.05442} {Establishing trustworthiness: Rethinking tasks and model evaluation}.

\bibitem[{Liu et~al.(2023)Liu, Zhang, and Liang}]{liu2023evaluating}
Nelson~F. Liu, Tianyi Zhang, and Percy Liang. 2023.
\newblock \href {http://arxiv.org/abs/2304.09848} {Evaluating verifiability in generative search engines}.

\bibitem[{Ma et~al.(2023)Ma, Zhang, Bian, Liu, Zhang, Zhao, Zhang, Fu, Hu, and Wu}]{DBLP:journals/corr/abs-2303-13217}
Huan Ma, Changqing Zhang, Yatao Bian, Lemao Liu, Zhirui Zhang, Peilin Zhao, Shu Zhang, Huazhu Fu, Qinghua Hu, and Bingzhe Wu. 2023.
\newblock \href {https://doi.org/10.48550/ARXIV.2303.13217} {Fairness-guided few-shot prompting for large language models}.
\newblock \emph{CoRR}, abs/2303.13217.

\bibitem[{Mei et~al.(2023)Mei, Levy, and Wang}]{mei2023foveate}
Alex Mei, Sharon Levy, and William~Yang Wang. 2023.
\newblock \href {http://arxiv.org/abs/2212.09667} {Foveate, attribute, and rationalize: Towards physically safe and trustworthy ai}.

\bibitem[{Min et~al.(2023)Min, Krishna, Lyu, Lewis, tau Yih, Koh, Iyyer, Zettlemoyer, and Hajishirzi}]{min2023factscore}
Sewon Min, Kalpesh Krishna, Xinxi Lyu, Mike Lewis, Wen tau Yih, Pang~Wei Koh, Mohit Iyyer, Luke Zettlemoyer, and Hannaneh Hajishirzi. 2023.
\newblock \href {http://arxiv.org/abs/2305.14251} {Factscore: Fine-grained atomic evaluation of factual precision in long form text generation}.

\bibitem[{Ni et~al.(2022)Ni, Qu, Lu, Dai, Hernandez~Abrego, Ma, Zhao, Luan, Hall, Chang, and Yang}]{ni-etal-2022-large}
Jianmo Ni, Chen Qu, Jing Lu, Zhuyun Dai, Gustavo Hernandez~Abrego, Ji~Ma, Vincent Zhao, Yi~Luan, Keith Hall, Ming-Wei Chang, and Yinfei Yang. 2022.
\newblock \href {https://doi.org/10.18653/v1/2022.emnlp-main.669} {Large dual encoders are generalizable retrievers}.
\newblock In \emph{Proceedings of the 2022 Conference on Empirical Methods in Natural Language Processing}, pages 9844--9855, Abu Dhabi, United Arab Emirates. Association for Computational Linguistics.

\bibitem[{Petroni et~al.(2022)Petroni, Broscheit, Piktus, Lewis, Izacard, Hosseini, Dwivedi-Yu, Lomeli, Schick, Mazaré, Joulin, Grave, and Riedel}]{petroni2022improving}
Fabio Petroni, Samuel Broscheit, Aleksandra Piktus, Patrick Lewis, Gautier Izacard, Lucas Hosseini, Jane Dwivedi-Yu, Maria Lomeli, Timo Schick, Pierre-Emmanuel Mazaré, Armand Joulin, Edouard Grave, and Sebastian Riedel. 2022.
\newblock \href {http://arxiv.org/abs/2207.06220} {Improving wikipedia verifiability with ai}.

\bibitem[{Phukan et~al.(2024)Phukan, Somasundaram, Saxena, Goswami, and Srinivasan}]{phukan2024peering}
Anirudh Phukan, Shwetha Somasundaram, Apoorv Saxena, Koustava Goswami, and Balaji~Vasan Srinivasan. 2024.
\newblock Peering into the mind of language models: An approach for attribution in contextual question answering.
\newblock \emph{arXiv preprint arXiv:2405.17980}.

\bibitem[{Pradeep et~al.(2021)Pradeep, Nogueira, and Lin}]{pradeep2021expandomonoduo}
Ronak Pradeep, Rodrigo Nogueira, and Jimmy Lin. 2021.
\newblock \href {http://arxiv.org/abs/2101.05667} {The expando-mono-duo design pattern for text ranking with pretrained sequence-to-sequence models}.

\bibitem[{Rashkin et~al.(2022)Rashkin, Nikolaev, Lamm, Aroyo, Collins, Das, Petrov, Tomar, Turc, and Reitter}]{rashkin2022measuring}
Hannah Rashkin, Vitaly Nikolaev, Matthew Lamm, Lora Aroyo, Michael Collins, Dipanjan Das, Slav Petrov, Gaurav~Singh Tomar, Iulia Turc, and David Reitter. 2022.
\newblock \href {http://arxiv.org/abs/2112.12870} {Measuring attribution in natural language generation models}.

\bibitem[{Sancheti et~al.(2024)Sancheti, Goswami, and Srinivasan}]{sancheti-etal-2024-post}
Abhilasha Sancheti, Koustava Goswami, and Balaji Srinivasan. 2024.
\newblock \href {https://doi.org/10.18653/v1/2024.starsem-1.4} {Post-hoc answer attribution for grounded and trustworthy long document comprehension: Task, insights, and challenges}.
\newblock In \emph{Proceedings of the 13th Joint Conference on Lexical and Computational Semantics (*SEM 2024)}, pages 49--57, Mexico City, Mexico. Association for Computational Linguistics.

\bibitem[{Sarti et~al.(2023{\natexlab{a}})Sarti, Chrupała, Nissim, and Bisazza}]{sarti2023quantifying}
Gabriele Sarti, Grzegorz Chrupała, Malvina Nissim, and Arianna Bisazza. 2023{\natexlab{a}}.
\newblock \href {http://arxiv.org/abs/2310.01188} {Quantifying the plausibility of context reliance in neural machine translation}.

\bibitem[{Sarti et~al.(2023{\natexlab{b}})Sarti, Feldhus, Sickert, and van~der Wal}]{sarti-etal-2023-inseq}
Gabriele Sarti, Nils Feldhus, Ludwig Sickert, and Oskar van~der Wal. 2023{\natexlab{b}}.
\newblock \href {https://doi.org/10.18653/v1/2023.acl-demo.40} {Inseq: An interpretability toolkit for sequence generation models}.
\newblock In \emph{Proceedings of the 61st Annual Meeting of the Association for Computational Linguistics (Volume 3: System Demonstrations)}, pages 421--435, Toronto, Canada. Association for Computational Linguistics.

\bibitem[{Toutanova et~al.(2003)Toutanova, Klein, Manning, and Singer}]{toutanova-etal-2003-feature}
Kristina Toutanova, Dan Klein, Christopher~D. Manning, and Yoram Singer. 2003.
\newblock \href {https://www.aclweb.org/anthology/N03-1033} {Feature-rich part-of-speech tagging with a cyclic dependency network}.
\newblock In \emph{Proceedings of the 2003 Human Language Technology Conference of the North {A}merican Chapter of the Association for Computational Linguistics}, pages 252--259.

\bibitem[{Touvron et~al.(2023)Touvron, Lavril, Izacard, Martinet, Lachaux, Lacroix, Rozière, Goyal, Hambro, Azhar, Rodriguez, Joulin, Grave, and Lample}]{touvron2023llama}
Hugo Touvron, Thibaut Lavril, Gautier Izacard, Xavier Martinet, Marie-Anne Lachaux, Timothée Lacroix, Baptiste Rozière, Naman Goyal, Eric Hambro, Faisal Azhar, Aurelien Rodriguez, Armand Joulin, Edouard Grave, and Guillaume Lample. 2023.
\newblock \href {http://arxiv.org/abs/2302.13971} {Llama: Open and efficient foundation language models}.

\bibitem[{Truong et~al.(2023)Truong, Baldwin, Verspoor, and Cohn}]{truong-etal-2023-language}
Thinh~Hung Truong, Timothy Baldwin, Karin Verspoor, and Trevor Cohn. 2023.
\newblock \href {https://doi.org/10.18653/v1/2023.starsem-1.10} {Language models are not naysayers: an analysis of language models on negation benchmarks}.
\newblock In \emph{Proceedings of the 12th Joint Conference on Lexical and Computational Semantics (*SEM 2023)}, pages 101--114, Toronto, Canada. Association for Computational Linguistics.

\bibitem[{Venkit et~al.(2024)Venkit, Chakravorti, Gupta, Biggs, Srinath, Goswami, Rajtmajer, and Wilson}]{venkit2024confidently}
Pranav~Narayanan Venkit, Tatiana Chakravorti, Vipul Gupta, Heidi Biggs, Mukund Srinath, Koustava Goswami, Sarah Rajtmajer, and Shomir Wilson. 2024.
\newblock " confidently nonsensical?'': A critical survey on the perspectives and challenges of'hallucinations' in nlp.
\newblock \emph{arXiv preprint arXiv:2404.07461}.

\bibitem[{Wang et~al.(2022)Wang, Mishra, Alipoormolabashi, Kordi, Mirzaei, Naik, Ashok, Dhanasekaran, Arunkumar, Stap, Pathak, Karamanolakis, Lai, Purohit, Mondal, Anderson, Kuznia, Doshi, Pal, Patel, Moradshahi, Parmar, Purohit, Varshney, Kaza, Verma, Puri, Karia, Doshi, Sampat, Mishra, Reddy~A, Patro, Dixit, and Shen}]{wang-etal-2022-super}
Yizhong Wang, Swaroop Mishra, Pegah Alipoormolabashi, Yeganeh Kordi, Amirreza Mirzaei, Atharva Naik, Arjun Ashok, Arut~Selvan Dhanasekaran, Anjana Arunkumar, David Stap, Eshaan Pathak, Giannis Karamanolakis, Haizhi Lai, Ishan Purohit, Ishani Mondal, Jacob Anderson, Kirby Kuznia, Krima Doshi, Kuntal~Kumar Pal, Maitreya Patel, Mehrad Moradshahi, Mihir Parmar, Mirali Purohit, Neeraj Varshney, Phani~Rohitha Kaza, Pulkit Verma, Ravsehaj~Singh Puri, Rushang Karia, Savan Doshi, Shailaja~Keyur Sampat, Siddhartha Mishra, Sujan Reddy~A, Sumanta Patro, Tanay Dixit, and Xudong Shen. 2022.
\newblock \href {https://doi.org/10.18653/v1/2022.emnlp-main.340} {Super-{N}atural{I}nstructions: Generalization via declarative instructions on 1600+ {NLP} tasks}.
\newblock In \emph{Proceedings of the 2022 Conference on Empirical Methods in Natural Language Processing}, pages 5085--5109, Abu Dhabi, United Arab Emirates. Association for Computational Linguistics.

\bibitem[{Wei et~al.(2023)Wei, Wang, Schuurmans, Bosma, Ichter, Xia, Chi, Le, and Zhou}]{wei2023chainofthought}
Jason Wei, Xuezhi Wang, Dale Schuurmans, Maarten Bosma, Brian Ichter, Fei Xia, Ed~Chi, Quoc Le, and Denny Zhou. 2023.
\newblock \href {http://arxiv.org/abs/2201.11903} {Chain-of-thought prompting elicits reasoning in large language models}.

\bibitem[{Zhao et~al.(2023)Zhao, Chen, Wang, Jiao, Long, Qin, Ding, Guo, Li, Li, and Joty}]{DBLP:conf/emnlp/ZhaoCWJLQDGLLJ23}
Ruochen Zhao, Hailin Chen, Weishi Wang, Fangkai Jiao, Do~Xuan Long, Chengwei Qin, Bosheng Ding, Xiaobao Guo, Minzhi Li, Xingxuan Li, and Shafiq Joty. 2023.
\newblock \href {https://aclanthology.org/2023.findings-emnlp.314} {Retrieving multimodal information for augmented generation: {A} survey}.
\newblock In \emph{Findings of the Association for Computational Linguistics: {EMNLP} 2023, Singapore, December 6-10, 2023}, pages 4736--4756. Association for Computational Linguistics.

\end{thebibliography}

\appendix
\clearpage
\section*{Appendix}

\section{Instruction Schema for CoG} 
\label{sec:decomp_schema}

\begin{mdframed}[backgroundcolor=gray!10, roundcorner=5pt]
You are a helpful assistant. You will be given a question and corresponding answer that is grounded to document. You need to break down the answer for a given question into information units. The answer is already split into sentences. Map each sentence from the answer to the corresponding information unit/ units. Give only those information units that are attributable to the grounded document. 

Instruction on what good information units are:
\begin{enumerate}
    \item Give information units that are relevant to the sentence.
    \item Information units should be meaningful.
    \item Break down information units at conjunctions.
    \item Information units should be co-referenced with respect to question.
    \item When the information units are put back together, it should convey the same information as the answer.
\end{enumerate}
Instruction on what bad information units are:
\begin{enumerate}
    \item Information units that convey duplicate information.
    \item Information units that are non statements.
    \item Information units that are not meaningful to the question.
    \item Information units that repeat facts present in the answer for introduction, conclusion or summary of an answer.
\end{enumerate}
\end{mdframed}

\pagebreak
\begin{mdframed}[backgroundcolor=gray!10, roundcorner=5pt]
\textbf{Examples:}

\textbf{QUESTION:}
\begin{quote}
    "Where was 'For You' by Rita Ora filmed?"
\end{quote}

\textbf{ANSWER:}
\begin{quote}
    [1:\{"The music video for `For You' by Liam Payne and Rita Ora was filmed at Oheka Castle on Long Island, off the coast of the eastern United States."\}]
\end{quote}

\textbf{GOOD ATOMIC FACTS:}

    \{"The song `For You' is performed by Liam Payne." : 1,
    "The song `For You' is performed by Rita Ora." : 1,
    "The music video for `For You' was filmed at Oheka Castle." : 1,
    "Oheka Castle is located on Long Island." : 1,
    "Long Island is off the coast of the eastern United States." : 1\}

\textbf{BAD ATOMIC FACTS:}

\{"The song is `For You'." : 1,
"Liam Payne and Rita Ora were filmed." : 1 \}

\end{mdframed}

\pagebreak

\section{Instruction Schema for LLM Attributor}
\label{sec:llm_att}
\begin{mdframed}[backgroundcolor=gray!10, roundcorner=5pt]

Given a question, information units relevant to the question and retrieved evidences, retrieve sentences from the evidence which support the information units. The sentences which support the information unit will be considered attributions to the information unit. The sentence should provide a reasoning to the information unit, question and answer.

Output a list of retrieved sentences. Output only a valid list and no other text. If no sentence is supported, return empty list []. Be precise in identifying sentences that support the information units by returning only highly relevant sentences. Return a list of length 0, 1 or 2. Do not return more. DO NOT PARAPHRASE THE SENTENCES FROM THE RETRIEVED EVIDENCES. OUTPUT EXACT SENTENCES IN THE LIST. Sort the returned list based on the relevance to the information unit. The highly relevant evidence should appear as the first element.

\textbf{OUTPUT FORMAT}: ["sentence 23", "sentence 34", "sentence 40"]

\textbf{QUESTION}: \texttt{\{question\}}

\textbf{ANSWER}: \texttt{\{answer\}}

\textbf{INFORMATION UNITS}: \texttt{\{iu\}}

\textbf{EVIDENCES}: \texttt{\{evidences\}}

Output a valid python list from now on

\textbf{OUTPUT LIST}: 
\end{mdframed}

\section{Dataset Statistics}
\label{sec:data_stat}
Table \ref{tab:stats} tabulates the statistics of the dataset used for evaluation.

\begin{table}[ht]
    \centering
    \scriptsize
    \begin{tabular}{l|c|c}
    \toprule
    \textbf{Dataset} & \textbf{Verifiability} & \textbf{QASPER} \\
    \midrule
    No. of Questions & 136 & 599 \\
    Avg. No. of source sentences & 130.03 & 517.49 \\
    Avg. No. of sentences per answer & 3.57 & 1.67 \\
    Avg. No. of words per answer & 72.96 & 14.8 \\
    Avg. No. of attributions per answer sentence & 1 & NIL \\
    Avg. No. of attributions per answer & 2.13 & 3.65 \\
    Avg. No. of answers per question & 2.75 & 1.47 \\
    \bottomrule
    \end{tabular}
    \caption{Dataset statistics reported on test sets.}
    \label{tab:stats}
\end{table}

\begin{table*}
    \centering
    \scalebox{0.66}{
\begin{tabular}{|c|c|c|c|c|c|}
\hline \multirow{3}{*}{\begin{tabular}{l} 
Question \\
Answer \\ \\
GT Attribution \\
\end{tabular}} & \multicolumn{5}{|c|}{ "paint cast iron" } \\
\hline & \multicolumn{5}{|c|}{\begin{tabular}{l} 
"To paint cast iron, you should first coat it with oil-based primer to create a smooth surface \\ and help the paint adhere. You can find cast iron paint on Amazon."
\end{tabular}} \\
\hline & CoG Decompositions & NIL-MonoT5 & CoG-MonoT5 & NIL-GPT4 & CoG-GPT4 \\
\hline \begin{tabular}{l} 
["Coat the cast \\ 
iron with oil- \\ 
based primer.', \\
"Priming the metal \\
creates a smooth \\
surface and will \\ 
help the paint adhere."]
\end{tabular} & \begin{tabular}{l} 
["To paint cast \\ 
iron, you should \\ 
first coat it with \\
oil-based primer.", \\ 
"The oil-based \\ 
primer helps create \\
a smooth surface \\
and help the paint \\
adhere."]
\end{tabular} & \begin{tabular}{l} 
["If you're working \\
with a smaller \\ 
piece of cast \\
iron, you can \\ 
wipe it down \\
with a damp rag, \\
instead.", \\
"Apply oil-based \\
paint to the \\
cast iron."]
\end{tabular} & \begin{tabular}{l} 
["Coat the cast \\
iron with oil- \\
based primer.",\\
"If you're working \\
with a smaller \\
piece of cast \\
iron, you can \\
wipe it down \\
with a damp \\
rag, instead."]
\end{tabular} & \begin{tabular}{l}
["Priming the metal \\
creates a smooth \\
surface and will \\
help the paint \\
adhere.",\\
"Apply oil-based \\
paint to the \\
cast iron.", \\
"Coat the cast \\
iron with oil\\
-based primer."]
\end{tabular} & \begin{tabular}{l} 
["Coat the cast iron \\
with oil-based \\
primer.", \\
"Priming the metal \\
creates a smooth \\
surface and will \\
help the paint \\
adhere."]
\end{tabular} \\
\hline [] & \begin{tabular}{l} 
["You can find \\
cast iron paint \\ 
on Amazon."]
\end{tabular} & \begin{tabular}{l} 
["Read on for \\
our complete guide \\
to painting cast \\
iron easily at home."]
\end{tabular} & \begin{tabular}{l} 
["Read on for \\
our complete guide \\
to painting cast \\
iron easily at home."]
\end{tabular} & [] & [] \\
\hline
\end{tabular}
}
\caption{Qualitative example of how decomposition affects retrieval based attributor and LLM based attributor. GT refers to ground truth. Each row depicts an answer part and respective decompositions and attributions for each method.}
    \label{tab:ret_llm}
\end{table*}

\section{Qualitative Example to compare retriever and LLM}
\label{sec:ret_llm}
Table \ref{tab:ret_llm} shows an example of how question in context of attribution enables LLMs to perform better attributions.

\section{Qualitative Examples for Ablation Studies}
\label{sec:abl}
Table \ref{tab:example} sets the ground by setting the context of the example for the qualitative ablation stuides in Table \ref{tab:ret_decomp_abl} and Table \ref{tab:llm_decomp_abl}.

\begin{table*}
    \centering
    \scalebox{0.66}{
\begin{tabular}{|c|c|}
\hline Question & "Which downstream tasks are used for evaluation in this paper?" \\
\hline Answer & \begin{tabular}{l} 
"Various sequence tagging tasks: Argument detection,\\
ACE entity and event detection, part-of-speech tagging, \\
CoNLL chunking, CoNLL named entity recognition, \\
GENIA bio-entity recognition, WNUT named entity \\
recognition. They also evaluate on Stanford \\
Sentiment Treebank, Penn TreeBank constituency parsing,\\
and Stanford Natural Language Inference."
\end{tabular} \\
\hline CoG IUs & \begin{tabular}{l} 
["Argument detection is used for evaluation in this paper.",\\
"ACE entity and event detection is used for evaluation in this paper.",\\
"CoNLL named entity recognition is used for evaluation",\\
"CoNLL chunking is used for evaluation.",\\
"WNUT named entity recognition is used for evaluation.",\\
"Part-of-speech tagging is used for evaluation in this paper.",\\
"GENIA bio-entity recognition is used for evaluation."]
\end{tabular} \\
\hline GT Attributions & \begin{tabular}{l} 
["GENIA NER: The Bio-Entity Recognition Task \\
at JNLPBA BIBREF9 annotated Medline abstracts \\
with information on bio-entities (like protein or DNA-names).", \\
"POS: We use the part-of-speech tags from \\
Universal Dependencies v. 1.3 for English \\
with the provided data splits.", \\
"We use the CoNLL 2003 NER model, the Stanford \\
Sentiment Treebank (SST-5) model, the \\
constituency parsing model for the Penn \\
TreeBank, and the Stanford Natural Language \\
Inference Corpus (SNLI) model.", \\
"ACE Entities/Events: ACE 2005 dataset \\
BIBREF8 consists of 599 annotated documents \\
from six different domains (newswire, broadcast\\
news, broadcast conversations, blogs,\\
forums, and speeches).", "We trained this \\
architecture for the following datasets:\\
Arguments: Argument component detection \\
(major claim, claim, premise) in 402\\
persuasive essays BIBREF7.", \\
"NER: CoNLL 2003 shared task on named entity recognition.", \\
"Chunking: CoNLL 2000 shared task dataset on chunking."]
\end{tabular} \\
\hline
\end{tabular}
}
\caption{Tabulation of question, answer, CoG decompositions and ground truth attributions for Table \ref{tab:ret_decomp_abl} and Table \ref{tab:llm_decomp_abl}}
\label{tab:example}
\end{table*}

\begin{table*}
    \centering
    \scalebox{0.66}{
\begin{tabular}{|c|r|r|}
\hline Evidence & NIL+MonoT5 & CoG+MonoT5 \\
\hline \begin{tabular}{c} 
"GENIA NER: The Bio-Entity Recognition Task at JNLPBA BIBREF9 annotated \\
Medline abstracts with information on bio-entities (like protein or DNA-names).",
\end{tabular} & -0.032 & -0.013\\
\hline "WNUT16: WNUT16 was a shared task on Named Entity Recognition over Twitter BIBREF10." & -0.043 &  -0.021\\
\hline
\end{tabular}
}
\caption{Example of retriever score getting affected while using answer part as information unit vs decomposed information unit. The example provided is for the ground truth evidence from Table \ref{tab:example}.}
    \label{tab:ret_decomp_abl}
\end{table*}

\begin{table*}
    \centering
    \scalebox{0.66}{
\begin{tabular}{|c|c|}
\hline NIL+GPT4 & CoG+GPT4 \\
\hline \begin{tabular}{l} 
"For the models included in AllenNLP, we observed a training \\
speed-up of 19-44\%, while improving the test performance in 3 \\
out of 5 datasets.", "'"The results for the second experiment, \\
that uses AllenNLP and ELMo embeddings in combination with \\
other input representations, are presented in the lower part of \\
Table $1 . "$, "Only for the GENIA dataset achieved the learned \\
weighted average a significantly better performance than using \\
the output of the second layer."
\end{tabular} & \begin{tabular}{l} 
"We trained this architecture for the following datasets: Arguments: Argument \\
component detection (major claim, claim, premise) in 402 persuasive essays \\
BIBREF7 .", "ACE Entities/Events: ACE 2005 dataset BIBREF8 consists of 599 \\
annotated documents from six different domains (newswire, broadcast news, \\
broadcast conversations, blogs, forums, and speeches).", "POS: We use the \\
part-of-speech tags from Universal Dependencies v. 1.3 for English with the \\
provided data splits.", "Chunking: CoNLL 2000 shared task dataset on \\
chunking.", "GENIA NER: The Bio-Entity Recognition Task at JNLPBA BIBREF9 \\
annotated Medline abstracts with information on bio-entities (like protein or \\
DNA-names).", "WNUT16: WNUT16 was a shared task on Named Entity \\
Recognition over Twitter BIBREF10 .", "We use the CoNLL 2003 NER model, the \\
Stanford Sentiment Treebank (SST-5) model, the constituency parsing model for \\
the Penn TreeBank, and the Stanford Natural Language Inference Corpus (SNLI) \\
model."
\end{tabular} \\
\hline
\end{tabular}
}
\caption{Example of LLM attributions getting affected while using answer part as information unit vs decomposed information unit.}
    \label{tab:llm_decomp_abl}
\end{table*}

\section{Human Survey}
\begin{figure*}[ht]
    \centering
    \includegraphics[width=0.8\linewidth]{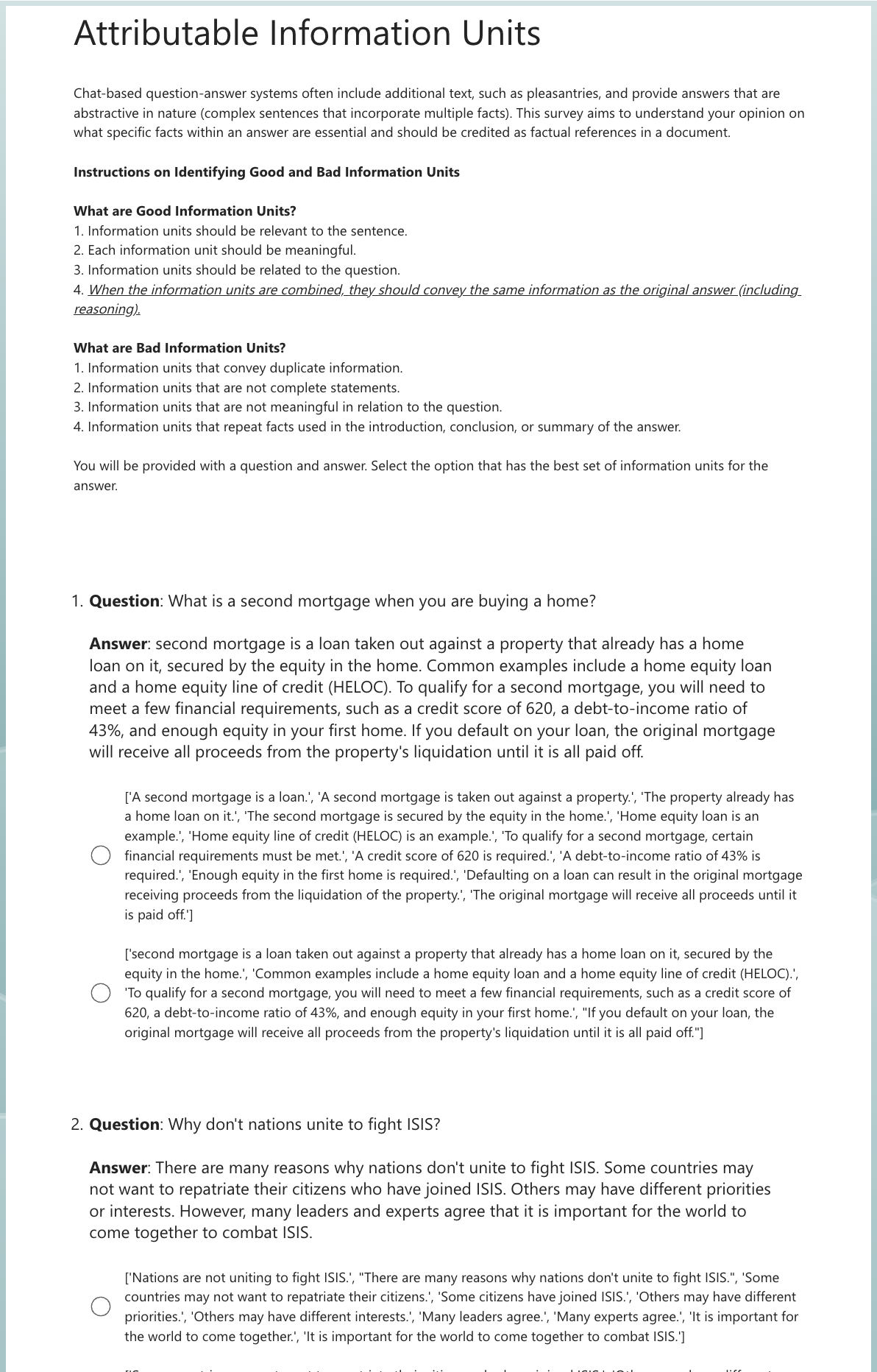} 
    \caption{Screenshot of Microsoft Forms used for survey.}
    \label{fig:survey}
\end{figure*}

Figure \ref{fig:survey} displays the survey format and the instructions provided to human annotators. Each step in the template clearly delineates guidelines for identifying good and bad atomic facts. The form included a radio button for evaluators to select which decomposition they believe best matches between the FActScore methodology and our approach. Notably, our definition of good atomic facts considers the relevance of the decomposition to the posed question.

\begin{figure*}[ht]
    \centering
    \begin{subfigure}[b]{0.8\linewidth}
        \centering
        \includegraphics[width=0.8\linewidth]{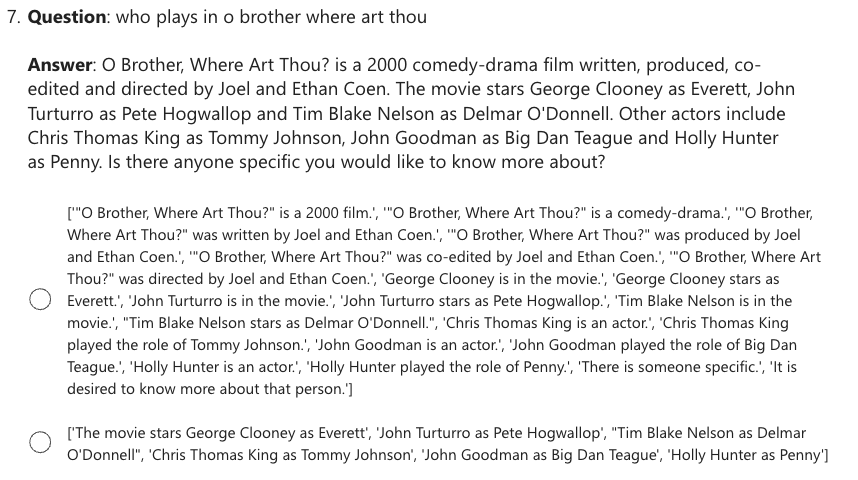}
        \caption{Example question and answer decomposition. First option shows FActScore decomposition and second option shows CoG decomposition.}
    \end{subfigure}
    \hfill
    \begin{subfigure}[b]{0.8\linewidth}
        \centering
        \includegraphics[width=0.8\linewidth]{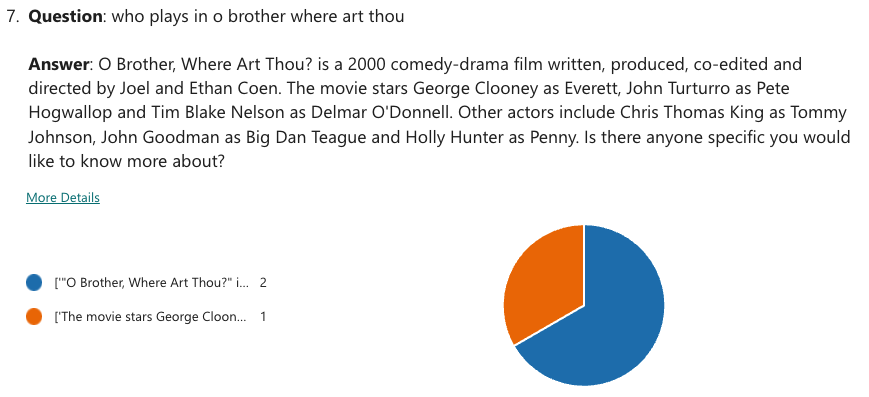}
        \caption{Case where human annotators preferred FActScore-based decomposition over CoG decomposition.}
    \end{subfigure}
    \caption{Human Annotation Error}
    \label{fig:survey_fail}
\end{figure*}

During this process, we encountered instances of human error in evaluation. Figure \ref{fig:survey_fail} illustrates a case where human annotators preferred the FActScore-based decomposition over the CoG decomposition. The most common reason for choosing FActScore over CoG was confusion among annotators about which information units were relevant to the question. For instance, Figure \ref{fig:survey_fail} relates to a question about the actor who starred in "O Brother, Where Art Thou." The answer includes introductory elements unrelated to the question, which ideally should not be attributed. The FActScore system fails to discriminate and attempts to generate facts for all sentences. In contrast, our system decomposes only those parts that are pertinent to both the question and the answer. This nuance was not captured during the human evaluation, leading to a preference for FActScore over our methodology, which is misleading. These instances significantly contributed to the error cases during evaluation.

\label{sec:survey}

\end{document}